\documentclass[conference]{IEEEtran}
\IEEEoverridecommandlockouts
\usepackage{cite}
\usepackage{amsmath,amssymb,amsfonts}
\usepackage{algorithmic}
\usepackage[pdftex]{graphicx}
\usepackage{textcomp}
\usepackage{xcolor}
\usepackage{comment}
\usepackage{multirow}
\usepackage{subfig}
\usepackage{url}

\def\BibTeX{{\rm B\kern-.05em{\sc i\kern-.025em b}\kern-.08em
    T\kern-.1667em\lower.7ex\hbox{E}\kern-.125emX}}
\begin{document}

\title{Hierarchical Group Sparse Regularization\\ for Deep Convolutional Neural Networks\\
\thanks{This work was partly supported by JSPS KAKENHI Grant Number 16K00239.}
}

\author{\IEEEauthorblockN{Kakeru Mitsuno}
\IEEEauthorblockA{\textit{School of Engineering} \\
\textit{Hiroshima University}\\
Higashi Hiroshima, Japan \\
mitsunokakeru@gmail.com
}
\and
\IEEEauthorblockN{Junichi Miyao}
\IEEEauthorblockA{\textit{Department of Information Engineering} \\
\textit{Hiroshima University}\\
Higashi Hiroshima, Japan \\
miyao@hiroshima-u.ac.jp
}
\and
\IEEEauthorblockN{Takio Kurita}
\IEEEauthorblockA{\textit{Department of Information Engineering} \\
\textit{Hiroshima University}\\
Higashi Hiroshima, Japan \\
tkurita@hiroshima-u.ac.jp}
}

\maketitle

\begin{abstract}
In a deep neural network (DNN), the number of the parameters is usually huge to get high learning performances. For that reason, it costs a lot of memory and substantial computational resources, and also causes overfitting.
It is known that some parameters are redundant and can be removed from the network without decreasing performance. 
Many sparse regularization criteria have been proposed to solve this problem. 
In a convolutional neural network (CNN), group sparse regularizations are often used to remove unnecessary subsets of the weights, such as filters or channels. 
When we apply a group sparse regularization for the weights connected to a neuron as a group, each convolution filter is not treated as a target group in the regularization. 
In this paper, we introduce the concept of hierarchical grouping to solve this problem, and we propose several hierarchical group sparse regularization criteria for CNNs. 
Our proposed the hierarchical group sparse regularization can treat the weight for the input-neuron or the output-neuron as a group and convolutional filter as a group in the same group to prune the unnecessary subsets of weights.
As a result, we can prune the weights more adequately depending on the structure of the network and the number of channels keeping high performance. 
In the experiment, we investigate the effectiveness of the proposed sparse regularizations through intensive comparison experiments on public datasets with several network architectures. 
Code is available on GitHub: \url{https://github.com/K-Mitsuno/hierarchical-group-sparse-regularization}
\end{abstract}


\section{Introduction}

Interest in methods of enforcing the network sparsity is increasing in the field of deep neural networks (DNN).
By making the network sparse, we can reduce the necessary computational resources, and improve the generalization performance of the trained network.

Tibshirami \cite{tibshirani1996regression} proposed the most simple non-structural sparse regularization lasso.
Zou and Hastie \cite{zou2005regularization} also proposed an elastic net that combined L2 regularization and L1 regularization as a weighted sum. 
Yuan and Lin \cite{yuan2006model} and Schmidt \cite{schmidt2010graphical} proposed group lasso regularization to neglect a group of parameters in the model. 
Kim and Xing \cite{Kim2010Tree} proposed tree-guided group lasso, which is based on group lasso, but groups are defined for a tree structure for a sparse multi-task regression.
Friedman et al. \cite{friedman2010note} and Simon et al. \cite{simon2013sparse} proposed sparse group lasso for linear regression, which combines L1 regularization and group lasso. 

Recently, some methods of pruning unnecessary weights of deep neural networks were proposed by many researchers.
Wen et al. \cite{wen2016learning} proposed a structured sparsity learning (SSL) method to regularize the structures of deep neural networks. 
SSL can learn a compact structure from a bigger DNN to reduce computation cost, obtain a hardware-friendly structured sparsity of DNN, and regularize the DNN to improve classification accuracy.

Alvarez and Salzmann \cite{alvarez2016learning} introduced an approach to automatically determine the number of neurons in each layer of a DNN during learning by using sparse group regularization.
This method can reduce the number of parameters by up to 80\% while retaining or even improving the network accuracy.
Scardapane et al. \cite{scardapane2017group} also proposed group sparse regularization for deep neural networks. 

Zhou et al. \cite{zhou2010exclusive} and Kong et al. \cite{kong2014exclusive} proposed exclusive lasso. 
Exclusive lasso introduces competition among variables in the same group.
Yoon and Hwang et al. \cite{yoon2017combined} proposed a combined group and exclusive sparsity (CGES) for deep neural networks. 
CGES enforces the network to be sparse and removes any redundancies in the features to fully utilize the capacity of the network.

Xu et al. \cite{xu20101} \cite{xu2012l_} \cite{zeng2014l_} proposed $L_{1/2}$ regularization. $L_{1/2}$ regularization can enforce the network to be more sparse than L1 regularization and much simpler than L0 regularization.
Fan et al. \cite{wu2014batch}  \cite{fan2014convergence}  applied $L_{1/2}$ regularization for sparsification of hidden layers of feed forward neural networks.
Li et al. \cite{li2018smooth} \cite{alemu2019group} proposed group $L_{1/2}$ regularization for feed forward neural networks. 

Li et al. \cite{li2019oicsr} proposed Out-In-Channel Sparse Regularization (OICSR) for compact deep neural networks. 
Ma et al. \cite{ma2019transformed} proposed non-convex integrated transformed L1 regularization for learning sparse deep neural networks. 
This method simultaneously promotes connection-level and neuron-level sparsity for DNNs.

%
%

These regularizations can make the weights of a deep neural network sparse at the individual weight level and the grouped weights level. 
In a convolutional neural network (CNN), we can consider the weights of a neuron as a group. 
However, we can also consider each convolution filter as a group.
To treat these groupings simultaneously, we have to introduce the concept of the hierarchical grouping.
In this paper, we propose several hierarchical group sparse regularization criteria for deep neural network pruning and evaluate the performance of regularization criteria through intensive comparison experiments.

\section{Related Works}

\begin{figure*}[htbp]
\centering
\subfloat[group lasso]{\includegraphics[scale=0.15]{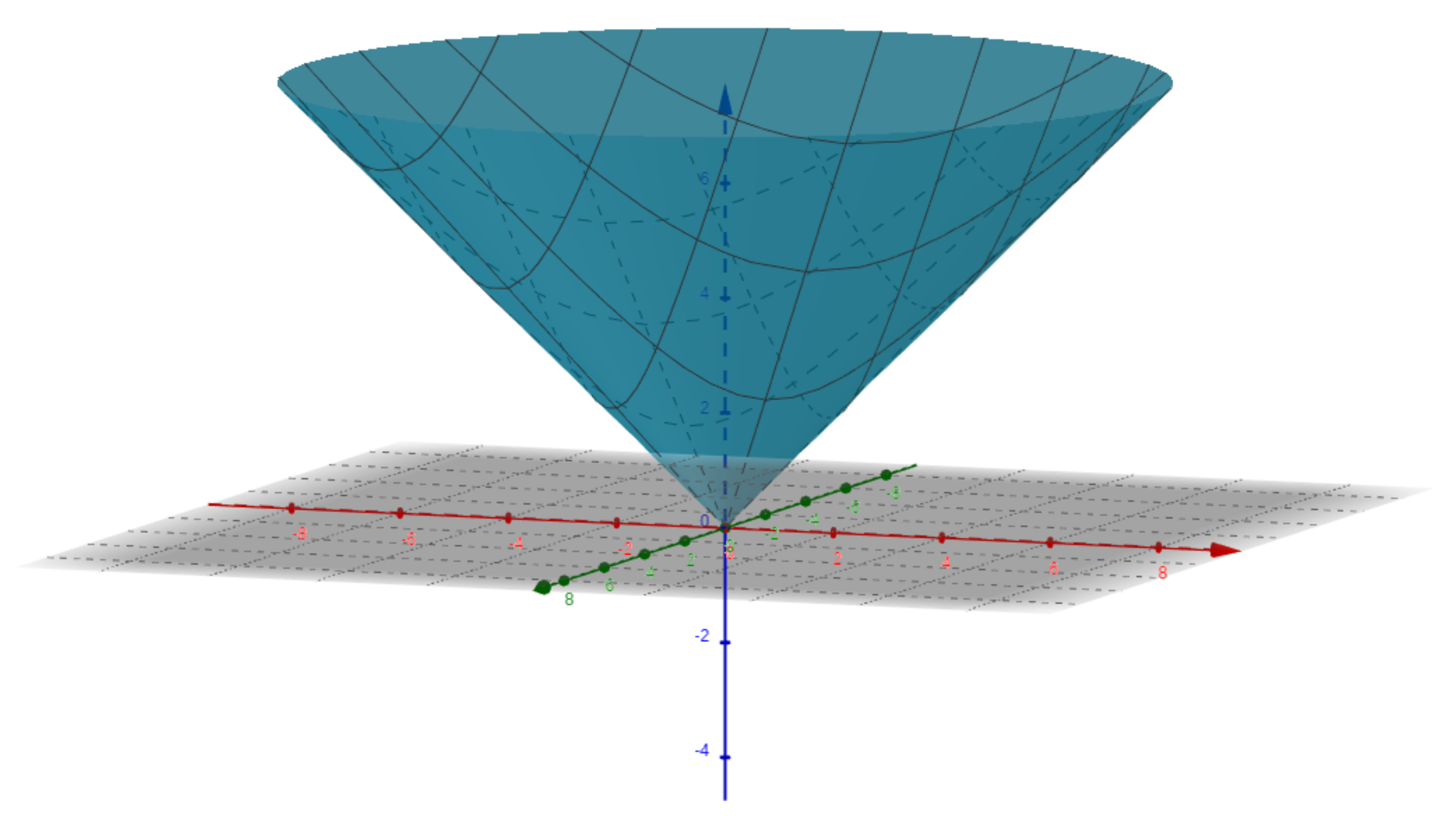}
\label{shape of group Lasso}}
\hfil
\subfloat[exclusive sparsity]{\includegraphics[scale=0.15]{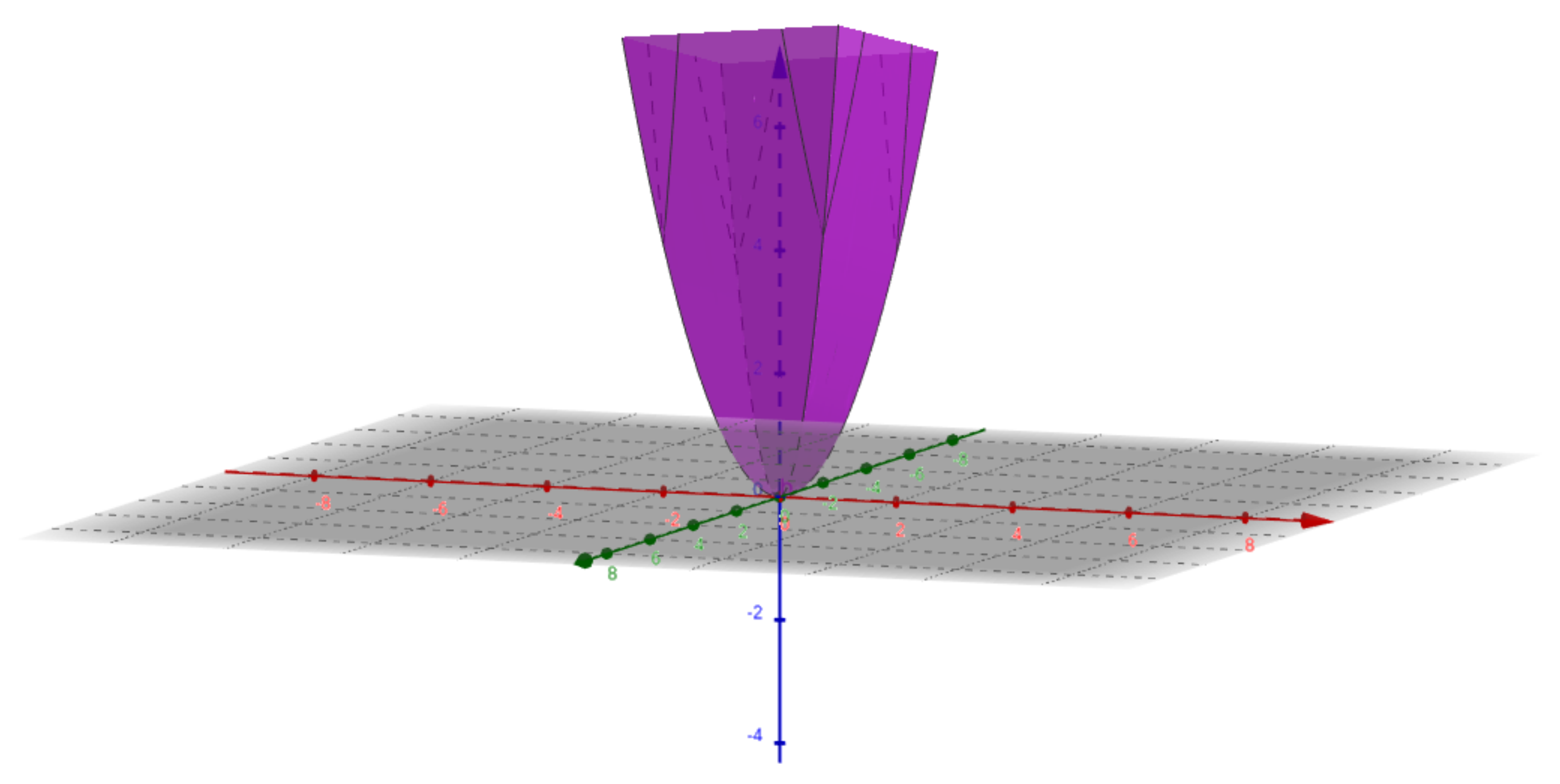}
\label{shape of exclusive sparsity}}
\hfil
\subfloat[group $L_{1/2}$ regularization]{\includegraphics[scale=0.15]{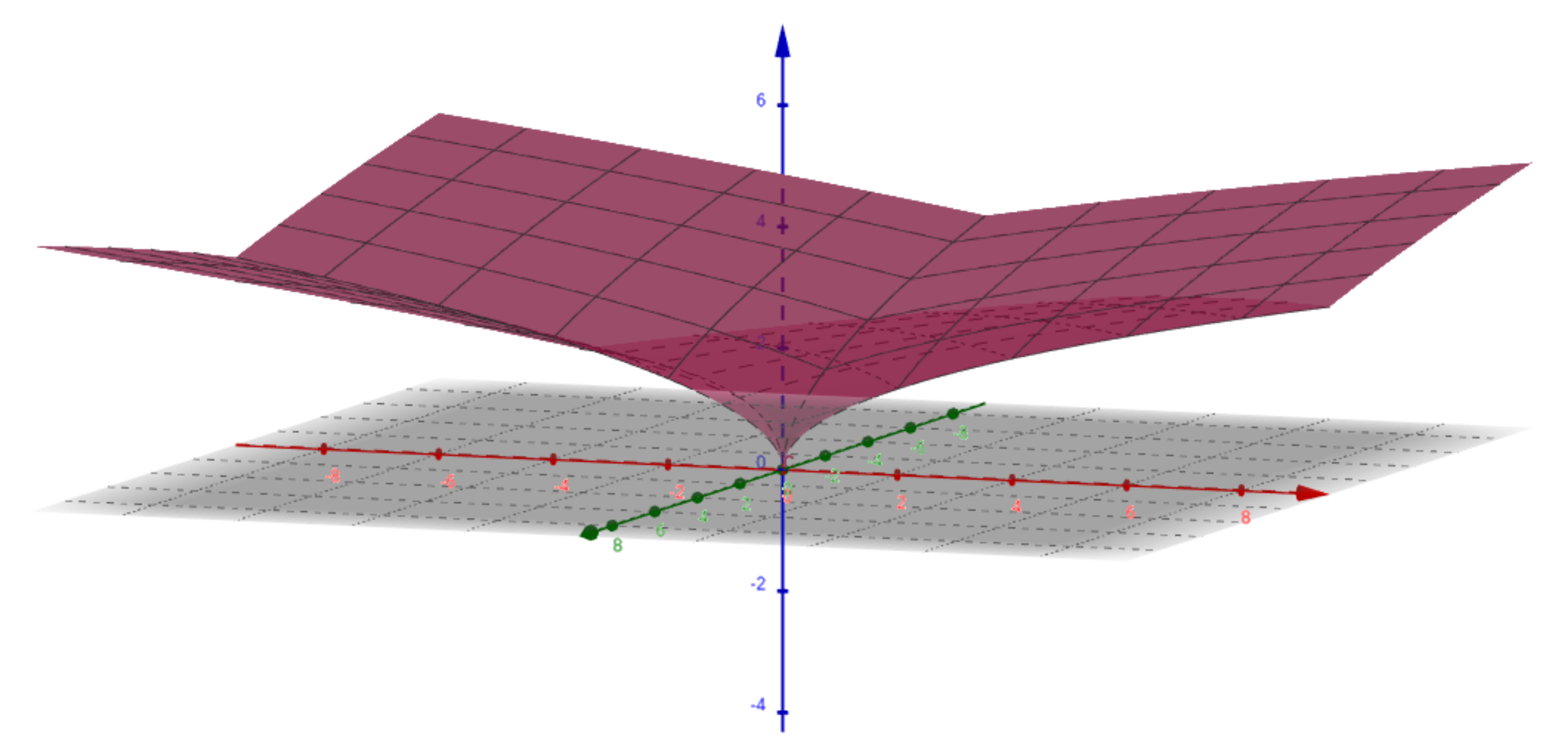}
\label{shape of group L1/2 regularization}}

\caption[]{The shape of three different regularization terms
\subref{shape of group Lasso} group lasso regularization
\subref{shape of exclusive sparsity} exclusive sparse regularization
\subref{shape of group L1/2 regularization} group $L_{1/2}$ regularization}
\label{shape of group Sparse Regularization}
\end{figure*}

In this section, we review previous works on weight pruning methods of deep neural networks in terms of the pruning criteria.

Assume that we have a training set with $N$ instances $D=\{(x_i,y_i)\}^{N}_{i=1}$, where $x_i \in \mathbb{R}^d$ is a $d$-dimensional input feature vector and $y_i \in \{1,\dots,K\}$ is a class label from one of the $K$ classes.
Then the objective function with sparse regularization for a deep neural network, especially for classification by convolutional neural networks (CNNs), can be represented as
\begin{equation}
\label{training objective}
J(W) = loss(W|D) + \lambda \sum^{L}_{l=1}R(W^l)
\end{equation}
where $loss(W|D)$ is the standard loss for the CNN, $W$ is the set of trainable weights for all $L$ layers in the CNN and $R(W^l)$ is the regularization term at $l^{th}$ layer for pruning the set of weights, $\{W^l\}$.
The parameter $\lambda$ is used to balance the loss and the pruning criterion.
If the $l^{th}$ layer is fully connected, we assume that the weight is given by $W^l \in \mathbb{R}^{oc_l\times ic_l}$,  where $oc_l$ and $ic_l$ are the dimensions of $W^l$ along the axes of out-channel and in-channel respectively. 
Also, we assume the weights as $W^l \in \mathbb{R}^{oc_l\times ic_l\times H_l \times W_l}$ when the $l^{th}$ layer is a convolutional layer, where $H_l$ and $ W_l$ are the height and width of the kernel respectively. 

The most often used sparse regularization is L2 regularization, defined as $\|W^l\|_2^2$. This regularization is often used in deep neural networks as weight decay to suppress over fitting.

Tibshirami \cite{tibshirani1996regression} proposed a simple non-structural sparse regularization as an L1 regularization for a linear model, which is defined as $\|W^l\|_1$. L1 regularization prevents overfitting by neglecting individual parameters in both convolution layers and fully connected layers. However, with L1 regularization, it is difficult to remove subsets of weights such as filters or channels in a CNN.



\subsection{Group Lasso Regularization}

Yuan and Lin \cite{yuan2006model} and Schmidt \cite{schmidt2010graphical} proposed group lasso regularization.
In order to reduce subsets of weights like filters or channels, it is necessary to treat the subsets as groups in the regularization criterion.
Yuan and Lin \cite{yuan2006model} and Schmidt \cite{schmidt2010graphical} proposed this regularization for a linear model that can treat sets of parameters as a group in the criterion. 
Group lasso forces subsets of unnecessary parameters to be simultaneously zero.
The regularization criterion of group lass is defined as
\begin{equation}
\label{group Lasso regularization}
R_{GL}(W^l) = \sum_{g \in G}\|W^l_g\|_2 = \sum_{g \in G} \sqrt{\sum_i {w^l_{g,i}}^2},
\end{equation}
where $g \in G$ is a group in the set of groups $G$, $W^l_g$ is the weight matrix or the weight vector for the group $g$ that is a sub matrix or sub vector in $W^l$ and $w^l_{g,i}$ is a weight with index $i$ in the group $g$. 
Group lasso introduces sparseness at the group level and can reduce the number of active neurons or active filters.  
Alvarez et al. \cite{alvarez2016learning} proposed an approach to automatically determine the number of neurons in each layer of a DNN during learning, and they showed that group lasso regularization could reduce the number of parameters and even improve network accuracy.
Wen et al. \cite{wen2016learning} proposed a structured sparsity learning (SSL) method to regularize the structures of deep neural networks by group lasso as structured sparse regularization. 
They introduced several structures of group lasso.

\subsection{Sparse Group Lasso Regularization}

Friedman et al. \cite{friedman2010note} and Simon et al. \cite{simon2013sparse} proposed sparse group lasso by combining L1 regularization and group lasso, applied to linear regression. 
Sparse group lasso forces parameters to be zero at both the group and the individual feature level.
Scardapane et al. \cite{scardapane2017group} proposed to use sparse group lasso for deep neural networks.
The criterion of the sparse group lasso is written as
\begin{equation}
\label{sparse group Lasso regularization}
R_{SGL}(W^l) = \alpha\sum_{g \in G} \|W^l_g\|_2 + (1-\alpha)\|W^l\|_1,
\end{equation}
where $\alpha$ is a balancing parameter to control strength of both group lasso and L1 regularization.
By this combination, unnecessary parameters in the network can be pruned at both the group level and the individual feature level.

\subsection{Exclusive Sparse Regularization}

Zhou et al. \cite{zhou2010exclusive} and Kong et al. \cite{kong2014exclusive} proposed exclusive lasso for multi-task feature selection. 
Exclusive lasso introduces competition among parameters in the same group and can prune neurons in neural networks.
It is also called exclusive sparsity and the regularization criterion is defined as
\begin{equation}
\label{exclusive Sparse Regularization}
R_{ES}(W^l) = \frac{1}{2}\sum_{g \in G}\|W^l_g\|^2_1 = \frac{1}{2}\sum_{g \in G} \left(\sum_i |w^l_{g,i}|\right)^2.
\end{equation}

\subsection{Combined Group and Exclusive Sparse Regularization}

Yoon and Hwang et al. \cite{yoon2017combined} proposed a pruning criterion called combined group and exclusive sparsity (CGES) for deep neural networks, which combines group lasso and exclusive sparse regularization. 
The authors claim that CGES can make the network sparse and also remove any redundancies among the features to fully utilize the capacity of the network.

\subsection{Group $L_{1/2}$ Regularization}

$L_{1/2}$ regularization, proposed by Xu et al. \cite{xu20101} \cite{xu2012l_} \cite{zeng2014l_}, can make the network to be more sparse than L1 regularization and much simpler than L0 regularization.
Fan et al. \cite{wu2014batch}  \cite{fan2014convergence}  applied $L_{1/2}$ regularization for pruning the neurons in the hidden layer of feedforward neural networks.
Li et al. \cite{li2018smooth} \cite{alemu2019group} also applied a group $L_{1/2}$ regularization for feedforward neural networks. 
$L_{1/2}$ regularization can make not only the redundant hidden nodes to be zero but also the redundant weights of the surviving hidden nodes of the neural networks to be zero.
In this paper, we define the criterion of the group $L_{1/2}$ regularization for deep neural network as 
\begin{equation}
\label{group L1/2 regularization}
R_{GL_{1/2}}(W^l) = \sum_{g \in G} \|W^l_g\|_{1}^{1/2} = \sum_{g \in G} \sqrt{\sum_i |w^l_{g,i}|}.
\end{equation}

\subsection{Out-In-Channel Sparse Regularization}

Li et al. \cite{li2019oicsr} proposed Out-In-Channel Sparse Regularization (OICSR) for compact deep neural networks. 
In OICSR, the correlations between successive layers are taken into consideration to keep the predictive power of the compact network.



\section{Proposed Method}

\subsection{Structured sparse regularization}

In this paper, we investigate the effectiveness of the structured sparse regularization criteria such as group lasso, exclusive sparsity, and group $L_{1/2}$ regularization for the convolutional neural network through intensive comparison experiments.
The definitions of these regularization criteria are shown as equations (\ref{group Lasso regularization}), (\ref{exclusive Sparse Regularization}), and (\ref{group L1/2 regularization}) respectively.
The visualization of these functions are shown in Fig.~\ref{shape of group Sparse Regularization}.

SSL proposed by Wen et al. \cite{wen2016learning} introduces various ways of grouping for structured sparse regularization.
In this paper, we also investigate the effectiveness of the ways of grouping through intensive comparison experiments.
In the following explanations, we will show the ways of grouping by using the criteria for group lasso, but we can also define the criteria for exclusive sparsity and group $L_{1/2}$ regularization.



\begin{figure}[htbp]
\centering
\subfloat[the filter-wise]{\includegraphics[scale=0.27]{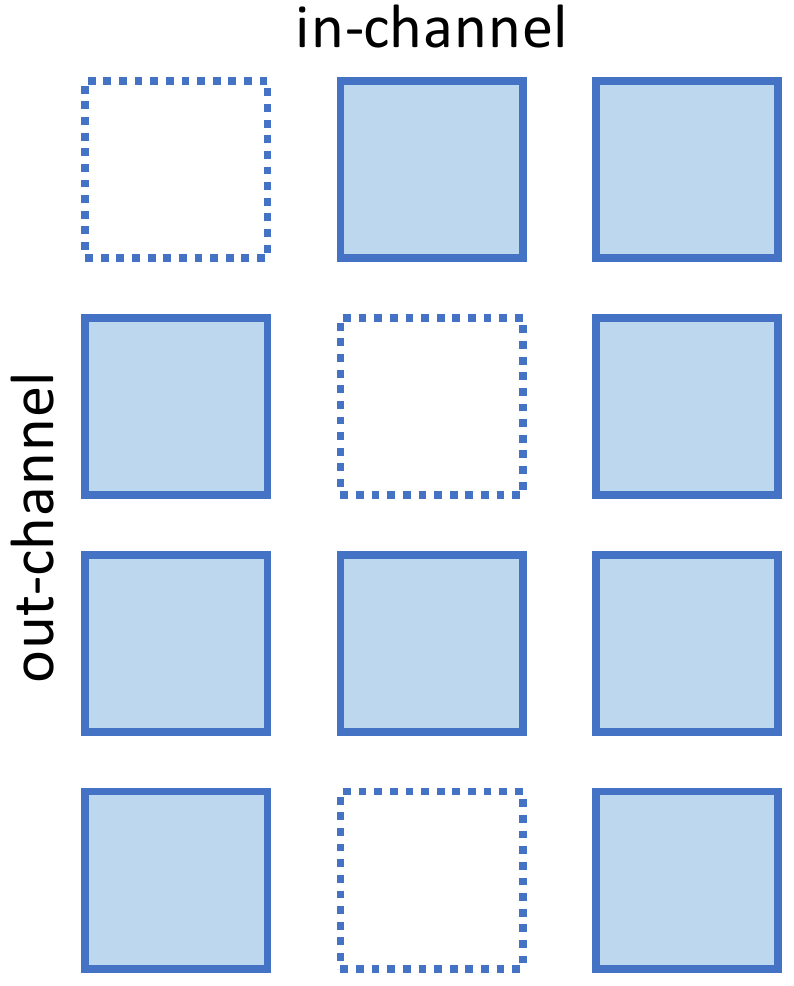}
\label{filter-wise grouping}}
\hfil
\subfloat[the neuron-wise]{\includegraphics[scale=0.27]{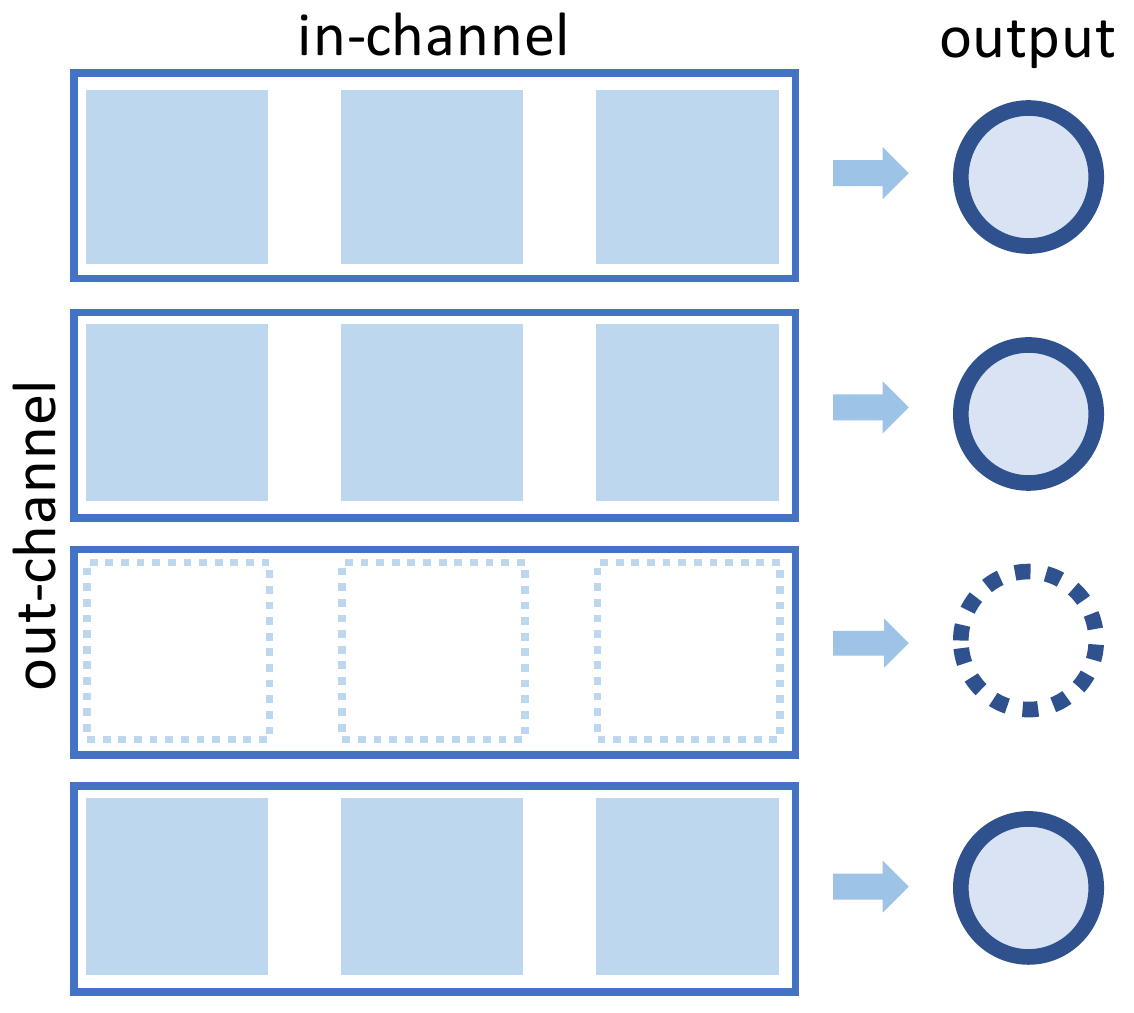}
\label{neuron-wise grouping}}
\hfil
\subfloat[the feature-wise]{\includegraphics[scale=0.27]{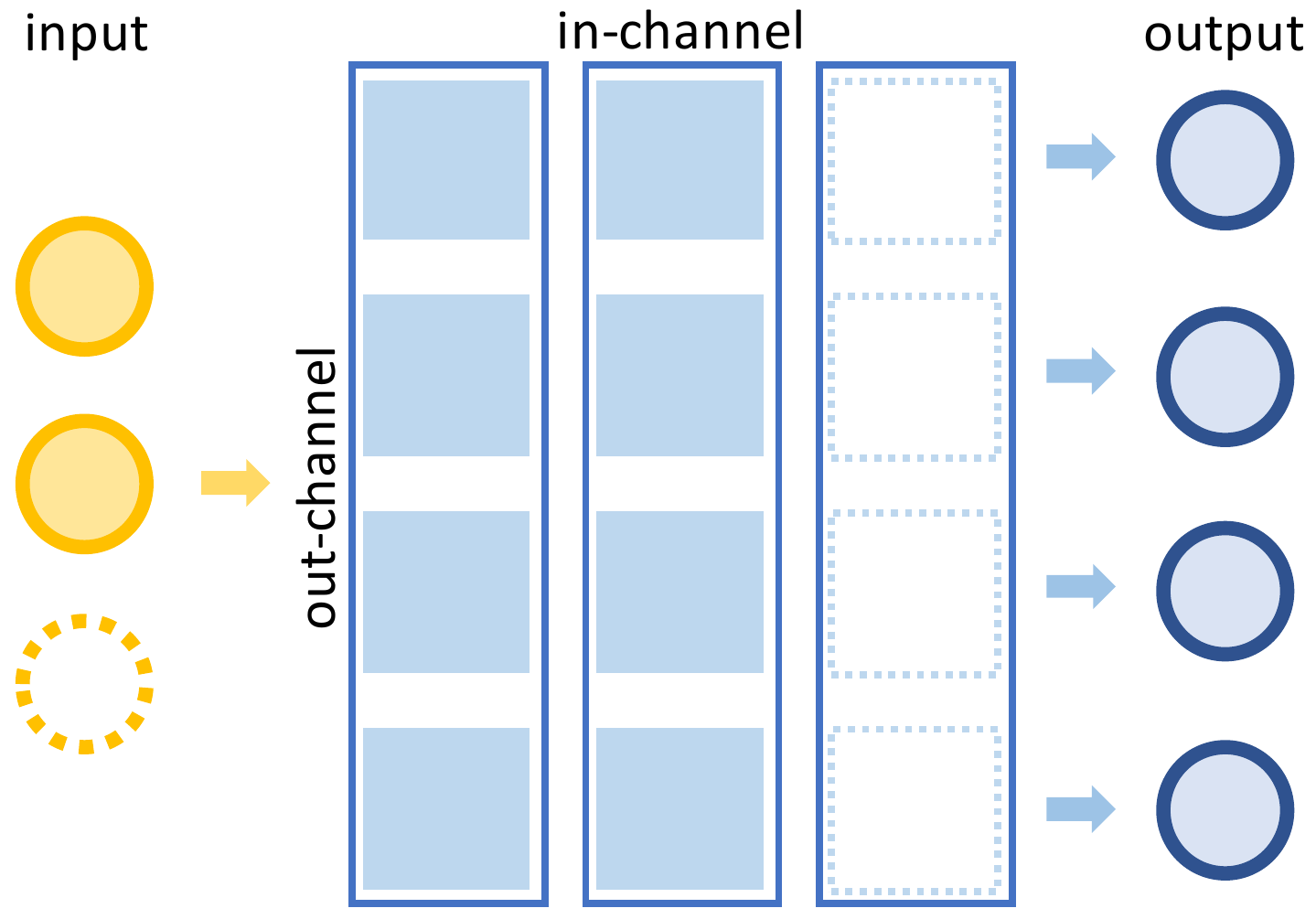}
\label{feature-wise grouping}}
\caption[]{The way of grouping for convolutional filters.
\subref{filter-wise grouping} Each filter is considered as a group. We call this grouping the filter-wise grouping. By this grouping, we can prune unnecessary filters.
\subref{neuron-wise grouping} The weights connected to a output neuron are consider as a group. We call this grouping the neuron-wise grouping. By this grouping, we can prune unnecessary output neurons.
\subref{feature-wise grouping} The weights connected to a input neuron are considered as a group. We call this grouping the feature-wise grouping. By this grouping, we can prune unnecessary the output channels in $(l-1)^{th}$ layer (the input channels in $(l)^{th}$ layer).}
\label{the way of grouping}
\end{figure}

In the case of a convolutional layer, we can consider three types of grouping for structured sparse regularization.
The way of grouping for a convolutional layer are shown in Fig.~\ref{the way of grouping}.
The first one is the filter-wise grouping which is defined as
\begin{equation}
\label{group lasso regularization for the filter-wise grouping in a convolutional layer}
R_{GL}(W^l) = \sum^{oc_l}_{i=1}\sum^{ic_l}_{j=1}\sqrt{\sum^{H_l}_{h=1}\sum^{W_l}_{w=1} {w^l_{i,j,h,w}}^2}.
\end{equation}
This criterion prunes unnecessary filters in the convolution layers.

The second one is the neuron-wise grouping which is defined as
\begin{equation}
\label{group lasso regularization for the neuron-wise grouping in convolutional layer}
R_{GL}(W^l) = \sum^{oc_l}_{i=1}\sqrt{\sum^{ic_l}_{j=1}\sum^{H_l}_{h=1}\sum^{W_l}_{w=1} {w^l_{i,j,h,w}}^2}.
\end{equation}
This criterion prunes unnecessary output neurons at each convolution layer.
As a result, the number of out-channels is reduced in each convolution layer. 

Last one is the feature-wise grouping which is defined as
\begin{equation}
\label{group Lasso regularization feature-wise grouping in convolutional layer}
R_{GL}(W^l) = \sum^{ic_l}_{j=1}\sqrt{\sum^{oc_l}_{i=1}\sum^{H_l}_{h=1}\sum^{W_l}_{w=1} {w^l_{i,j,h,w}}^2}.
\end{equation}
This criterion prunes unnecessary input neurons of the convolutional layer.
As a result, we can remove unnecessary out-channel in $(l-1)^{th}$ layer by making the unnecessary input neurons in $l^{th}$ layer zero.

\subsection{Hierarchical Group Sparse Regularization}

In the fully connected layer, weights are not structured, and we can apply sparse regularization to prune unnecessary input neurons or output neurons. 

On the other hand, the weights of the convolutional layer are structured as convolution filters, and there are three types of grouping, namely the filter-wise grouping, the neuron-wise grouping and the feature-wise grouping.


However, mutual interaction between filters in the group is not taken into account in the neuron-wise grouping or the feature-wise grouping.
To introduce such interactions in the sparse regularization criterion, we propose hierarchical group sparse regularization.

There are several possibilities to define the hierarchical interactions between filters in the group for structured sparse regularization.
In this paper, we consider two ways of the integration, namely the square root of the sub-groups and the square of the sub-groups. 
Thus, we can propose a set of hierarchical group sparse regularization criteria using group lasso regularization, exclusive sparsity, and group $L_{1/2}$ regularization based on the neuron-wise grouping or the feature-wise grouping. 
In the following, we explain the hierarchical group sparse regularization criteria based on the feature-wise grouping.
However, we can easily define the hierarchical group sparse regularization criteria based on the neuron-wise grouping.

The hierarchical group lasso regularization criterion based on the feature-wise groupings is defined as
\begin{equation}
\label{hierarchical group Lasso regularization square root}
R_{HSQRT-GL}(W^l) = \sum^{ic_l}_{j=1}\sqrt{\sum^{oc_l}_{i=1}\sqrt{\sum^{H_l}_{h=1}\sum^{W_l}_{w=1} {w^l_{i,j,h,w}}^2}}.
\end{equation}
In this criterion, the square root of the sub-groups (the feature-wise groupings) are taken to defined the sparse regularization criterion. 

The hierarchical group lasso regularization criterion based on the feature-wise groupings is also defined by taking the square of the sub-groups as
\begin{equation}
\label{hierarchical group Lasso regularization square}
R_{HSQ-GL}(W^l) =
\sum^{ic_l}_{j=1}\left( \sum^{oc_l}_{i=1}\sqrt{\sum^{H_l}_{h=1}\sum^{W_l}_{w=1} {w^l_{i,j,h,w}}^2}\right)^2.
\end{equation}
It is expected that these hierarchical group lasso criteria can prune unnecessary output neurons and input neurons simultaneously.

Similarly, the hierarchical exclusive sparse regularization criterion is define as
\begin{equation}
\label{hierarchical exclusive Sparse Regularization square root}
R_{HSQRT-ES}(W^l) = \sum^{ic_l}_{j=1}\sqrt{\sum^{oc_l}_{i=1}\left(\sum^{H_l}_{h=1}\sum^{W_l}_{w=1} |w^l_{i,j,h,w}|\right)^2}
\end{equation}
and
\begin{equation}
\label{hierarchical exclusive Sparse Regularization square}
R_{HSQ-ES}(W^l) =
\sum^{ic_l}_{j=1}\left( \sum^{oc_l}_{i=1}\left(\sum^{H_l}_{h=1}\sum^{W_l}_{w=1} |w^l_{i,j,h,w}|\right)^2\right)^2
\end{equation}

The hierarchical group $L_{1/2}$ regularization criterion is also defined as
\begin{equation}
\label{hierarchical group L1/2 regularization square root}
R_{HSQRT-GL_{1/2}}(W^l) = \sum^{ic_l}_{j=1}\sqrt{\sum^{oc_l}_{i=1}\sqrt{\sum^{H_l}_{h=1}\sum^{W_l}_{w=1} |w^l_{i,j,h,w}|}}
\end{equation}
and
\begin{equation}
\label{hierarchical group L1/2 regularization square}
R_{HSQ-GL_{1/2}}(W^l) =
\sum^{ic_l}_{j=1}\left( \sum^{oc_l}_{i=1}\sqrt{\sum^{H_l}_{h=1}\sum^{W_l}_{w=1} |w^l_{i,j,h,w}|}\right)^2.
\end{equation}


It is also possible to combine the L1 regularization with the hierarchical group sparse regularization criteria in order to prune unnecessary individual weights.

%
%

In the next section, we investigate the effectiveness of the each criterion through intensive comparison experiments.

\section{Experiments}

\subsection{The Sparse Regularization Criteria}

We have performed experiments with the convolutional neural network to compare the effectiveness of the sparse regularization criteria explained in this paper. 
They are summarized in the table \ref{Summary of sparse regularization term}.
The regularization is applied to the weights except for the bias term in all convolutional layers.

\begin{table}[htbp]
\centering
\caption{Summary of the sparse regularization criteria}
\label{Summary of sparse regularization term}
\begin{tabular}{c|l}\hline
abbreviation & sparse regularization criteria \\ \hline
L2 & L2 regularization \\
L1 & L1 regularization\cite{tibshirani1996regression} \\
GL & Group lasso regularization \cite{yuan2006model}\cite{schmidt2010graphical}\\
ES & Exclusive sparse regularization\cite{zhou2010exclusive}\cite{kong2014exclusive} \\
GL$_{1/2}$ & Group $L_{1/2}$ regularization\cite{li2018smooth}\cite{alemu2019group} \\
SGL & Sparse group lasso regularization\cite{scardapane2017group} \\
SGL$_{1/2}$ & Combined GL$_{1/2}$ and L1 \\
CGES & CGES regularization\cite{yoon2017combined} \\
OICSR-GL & Combined OICSR and GL\cite{li2019oicsr} \\ \hline
HSQRT-GL & Hierarchical square rooted GL \\
HSQ-GL & Hierarchical squared GL \\
HSQRT-ES & Hierarchical square rooted ES \\
HSQ-ES & Hierarchical squared ES \\
HSQRT-GL$_{1/2}$ & Hierarchical square rooted GL$_{1/2}$ \\
HSQ-GL$_{1/2}$ & Hierarchical squared GL$_{1/2}$ \\
SHSQRT-GL$_{1/2}$ & Combined HSQRT-GL$_{1/2}$ and L1 \\
SHSQ-GL$_{1/2}$ & Combined HSQ-GL$_{1/2}$ and L1 \\
\hline
\end{tabular}
\end{table}


\subsection{Networks and Datasets}



To confirm the robustness to the variations of the characteristics of the data, we have performed experiments using five datasets MNIST, Fashion-MNIST, CIFAR-10, CIFAR-100, and STL-10.
A simple CNN, AlexNet\cite{krizhevsky2012imagenet}, ResNet\cite{he2016deep} and VGG nets\cite{simonyan2014very} are used as the base network and they are trained from the scratch. 
The number of channels of the network at each layer is adjusted to prevent overfitting, depending on each dataset.

MNIST contains 70,000 grayscale images of handwritten digits. 
So the number of classes is 10.
The size of the image is $28 \times 28$ pixels. 
They are divided into 60,000 training images and 10,000 testing images. The simple CNN with two convolutional layers and two fully connected layers is trained by using the training images of MNIST dataset.

Fashion-MNIST contains 70,000 grayscale images of ten different fashion items. 
The size of the image is $28 \times 28$ pixels. 
They are divided into  60,000 training images and 10,000 testing images.
Similar to the MNIST, the simple CNN with two convolutional layers and three fully connected layers is trained by using the training images of Fashion-MNIST datasets.

CIFAR-10 contains 60,000 color images of ten different animals and vehicles.
The size of the image is $32 \times 32$ pixels. 
They are divided into 50,000 training images and 10,000 testing images. 
For CIFAR-10, we trained AlexNet with 5 convolutional layers and three fully connected layers and ResNet18 with 17 convolutional layers and one fully-connected layer with batch normalization layers.
The number of channels of AlexNet is set to [ 8, 16, 32, 16, 16, 128, 128, 10] and [ 8, 8, 8, 8, 8, 16, 16, 16, 16, 32, 32, 32, 32, 64, 64, 64, 64, 10] for ResNet18.

CIFAR-100 contains 60,000 color images of 100 different categories. 
The size of the image is $32 \times 32$ pixels. 
They are divided into 50,000 training images and 10,000 testing images.
For CIFAR-100, we trained VGG11bn, that has eight convolutional layers and three fully connected layers with batch normalization layers.
The number of channels is set to [ 16, 32, 64, 64, 128, 128, 128, 128, 128, 128, 10].

STL-10 contains 13,000 color images of animals and vehicles.
(airplane, bird, car, cat, deer, dog, horse, monkey, ship, truck). 
The size of the image is $96 \times 96$ pixels. 
They are divided into 5,000 training images and 8,000 testing images. 
For STL-10 dataset, we trained AlexNet with five convolutional layers and three fully connected layers and ResNet18 with 17 convolutional layers and one fully-connected layer with batch normalization layers.
The number of channels is set to  [ 16, 32, 64, 32, 32, 512, 512, 10] for AlexNet, and  [ 8, 8, 8, 8, 8, 16, 16, 16, 16, 32, 32, 32, 32, 64, 64, 64, 64, 10] for ResNet18.

\subsection{Experimental Setting}

All the base networks are trained by using SGD optimizer with a momentum of $0.9$.
Also, we used the weight decay with the strength of $10^{-4}$ to prevent overfitting. 
For MNIST and Fashion-MNIST, the networks are trained for $30$ epochs with the sparse regularization using a mini-batch size $256$. 
For CIFAR-10 and CIFAR-100, we trained the networks for $100$ epochs with the sparse regularization using a mini-batch size $128$. 
For STL-10, the network is trained for $100$ epochs with the sparse regularization using a mini-batch size $64$. 

The hyper-parameter $\lambda$, which balances the cross-entropy loss and the sparse regularization criterion, is experimentally determined by grid search in the range from $10^{-1} $ to $10^{-6}$.
For SGL, SGL$_{1/2}$, SHSQRT-GL$_{1/2}$, SHSQ-GL$_{1/2}$ and OICSR-GL, we set the parameter $\alpha$, which balances the L1 regularization and the group sparse regularization criterion, to be $0.5$. 
Also, we set $m = 0.8$ for CGES.

\subsection{Preliminary Experiments \\using MNIST and Fashion-MNIST datasets}

\begin{figure*}[htbp]
\centering
\subfloat[Filters of trained network without regularization]{\includegraphics[scale=0.33]{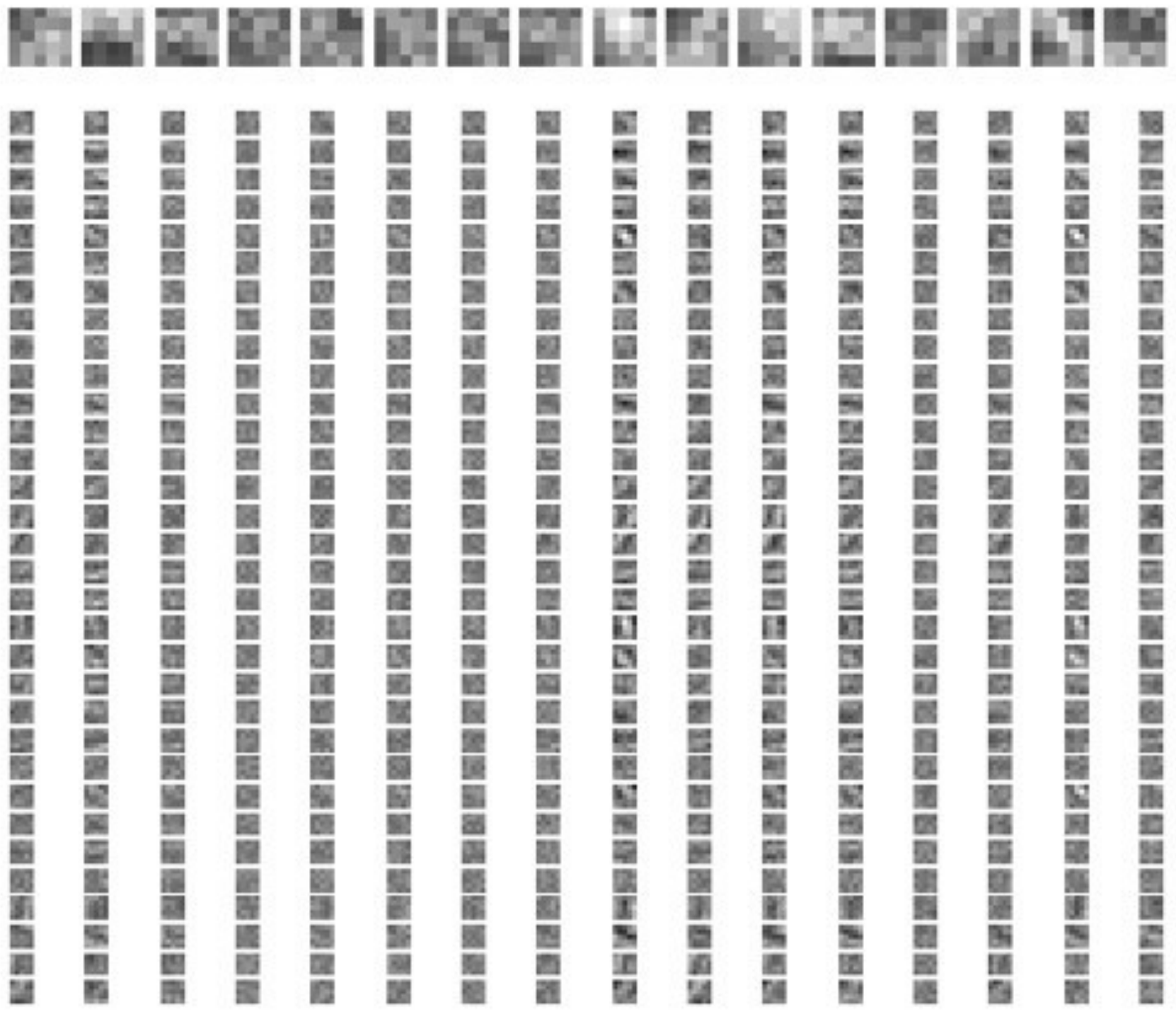}
\label{weights of convolution in baseline network}}
\hfill
\subfloat[Filters of trained network with neuron-wise HSQRT-GL]{\includegraphics[scale=0.33]{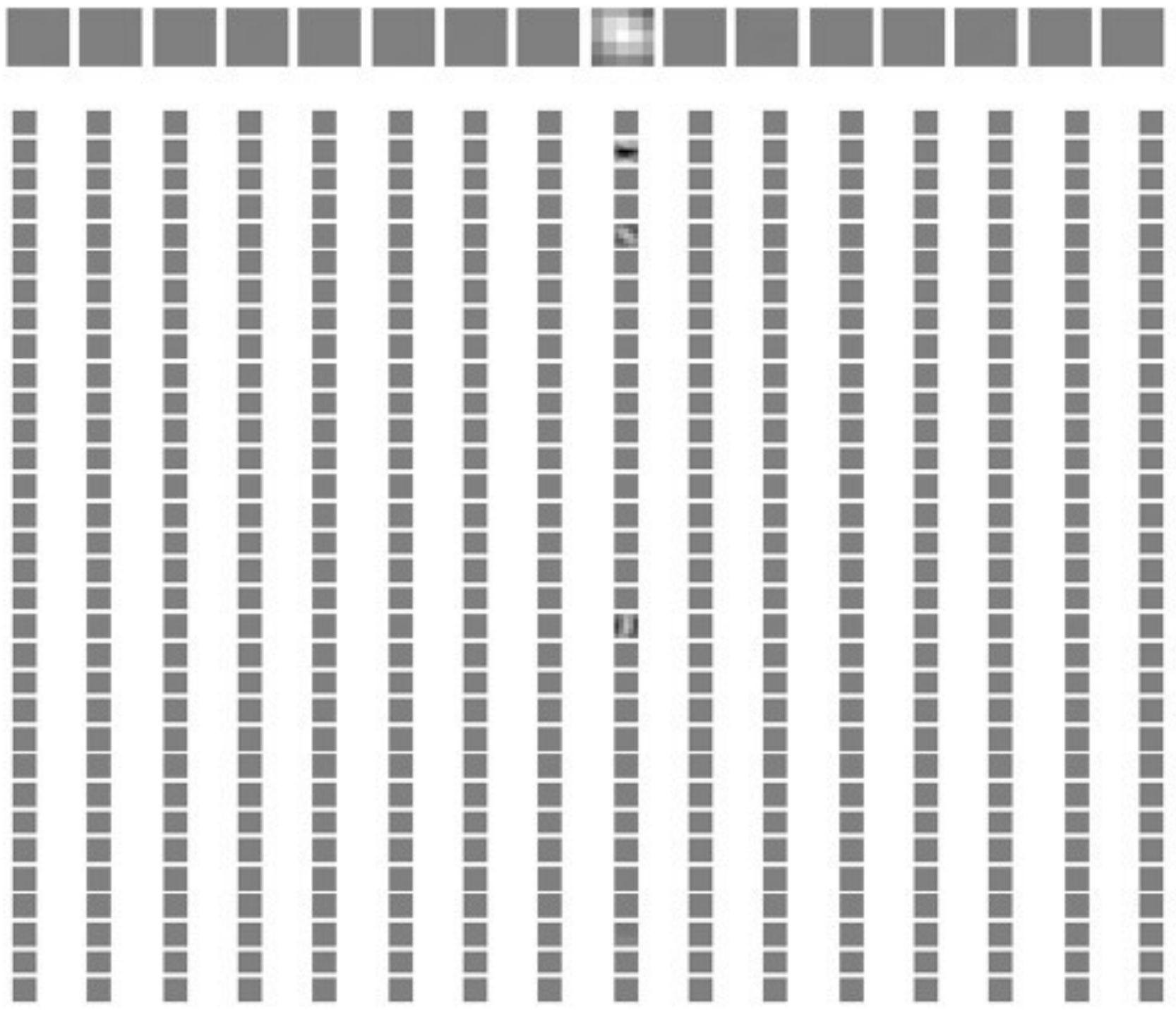}
\label{weights of convolution in trained network with HSQRT-GL_neuron}}
\hfill
\subfloat[Filters of trained network with feature-wise HSQRT-GL]{\includegraphics[scale=0.33]{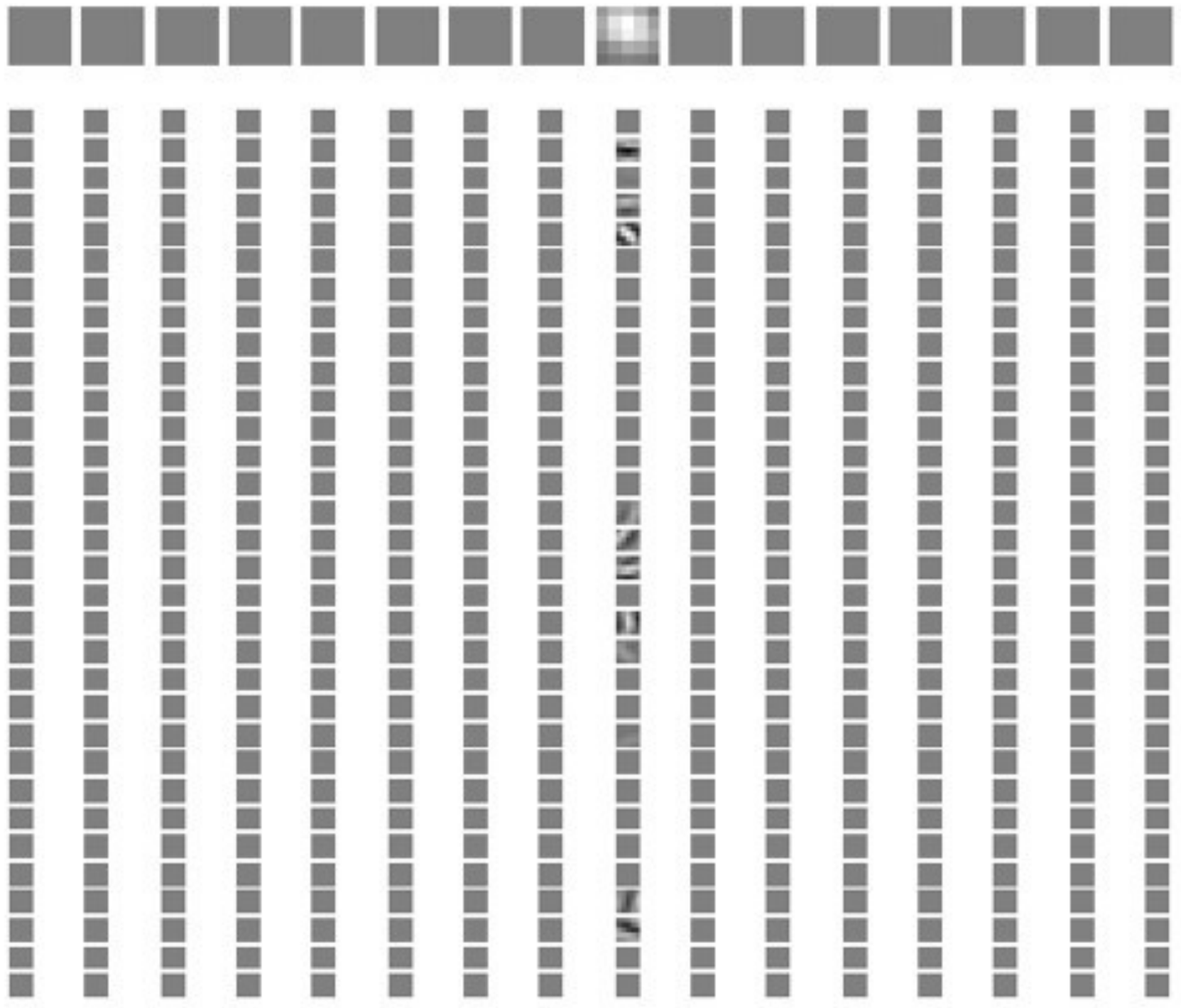}
\label{weights of convolution in trained network with HSQRT-GL_feature}}

\caption[]{Visualization of the $1_{st}$ convolutional layer filer (showed at above) and $2_{nd}$ convolutional layer filter for a input neuron (showed at below) from the network trained on MNIST dataset. The numbers of channels of each layer are [ 16, 32, 128, 10]. All filters are shown in $1_{st}$ convolutional layer and $2_{nd}$ convolutional layer.
\subref{weights of convolution in baseline network} Filters of trained network without regularization. Sparsity of the filters at the $1_{st}$ convolutional layer and $2_{nd}$ convolutional layer are 00.25\%, 1.86\% respectively.
\subref{weights of convolution in trained network with HSQRT-GL_neuron} Filters of trained network with neuron-wise HSQRT-GL. Sparsity of the filters at the $1_{st}$ convolutional layer and $2_{nd}$ convolutional layer are 63.50\%, 99.22\% respectively.
\subref{weights of convolution in trained network with HSQRT-GL_feature} Filters of trained network with feature-wise HSQRT-GL. Sparsity of the filters at the $1_{st}$ convolutional layer and $2_{nd}$ convolutional layer are 93.75\%, 97.75\% respectively.}
\label{weight after training}
\end{figure*}

\begin{table}[htbp]
\centering
\caption{Accuracy and sparsity with simple CNN on MNIST and Fashion-MNIST datasets. Top 2 sparsity are shown in boldface.}
\label{tab_SimpleNet_on_MNIST_FashionMNIST}
\begin{tabular}{c||cc||cc}\hline
Dataset & \multicolumn{2}{c||}{MNIST} & \multicolumn{2}{c}{Fashion-MNIST} \\ \hline
Method & Accuracy & Sparsity & Accuracy & Sparsity\\ \hline \hline
L2(Baseline) & 99.03\% & 1.81\% & 87.90\% & 1.73\% \\ 
L1 & 99.18\% & 39.29\% & 89.15\% & 95.39\% \\ 
GL  & 99.20\% & 1.82\% & 89.50\% & 68.66\% \\ 
ES & 99.14\% & 52.98\%  & 88.32\% & 98.76\% \\ 
GL$_{1/2}$ & 99.16\% & 50.33\% & 89.10\% & 96.61\% \\ 
SGL & 99.22\% & 19.62\% & 88.38\% & 99.20\% \\ 
SGL$_{1/2}$ & 99.24\% & 21.70\% & 88.26\% & 99.45\% \\ 
CGES & 99.05\% & 58.94\% & 89.69\% & 62.42\% \\ 
OICSR-GL & 99.24\% & 1.89\% & 89.50\% & 67.21\% \\  \hline
HSQRT-GL & 99.22\% & 2.20\% & 89.40\% & 87.95\% \\ 
HSQ-GL & 99.17\% & 20.55\% & 88.86\% & 86.05\% \\ 
HSQRT-ES &  99.09\% & 58.80\% & 88.78\% & 96.23\% \\ 
HSQ-ES & 99.10\% & 19.75\% & 88.05\% & 80.98\% \\ 
HSQRT-GL$_{1/2}$ & 99.25\% & 26.83\% & 87.97\% & \textbf{99.67\%} \\ 
HSQ-GL$_{1/2}$ & 99.13\% & \textbf{77.60\%} & 88.61\% & \textbf{99.54\%} \\ 
SHSQRT-GL$_{1/2}$ & 99.20\% & 21.09\% & 88.39\% & 99.43\% \\ 
SHSQ-GL$_{1/2}$ & 99.04\% & \textbf{94.08\%} & 88.82\% & 99.38\% \\ 
\hline
\end{tabular}
\end{table}

At first, we have performed preliminary experiments to investigate the effectiveness of the proposed hierarchical group sparse regularization for the simple CNN using MNIST and Fashion-MNIST datasets, which include gray-scale images.

Results of the simple CNN for MNIST and Fashion-MNIST with the hierarchical group sparse regularizations and the other sparse regularizations are shown in Tab.~\ref{tab_SimpleNet_on_MNIST_FashionMNIST}.
The ratio of the zero weights is calculated by assuming the weights whose absolute value is less than $10^{-3}$ are zero to evaluate the sparsity of the trained network.

From this table, it is noticed that all the test accuracies are higher than the baseline (L2) after the sparse regularizations are introduced.
For the MNIST dataset, the sparse regularizations SHSQ-GL$_{1/2}$ and HSQ-GL$_{1/2}$ achieved the sparsity of $94.08$\% and $77.60$\%.
For the Fashion-MNIST dataset, the sparse regularizations HSQRT-GL$_{1/2}$, HSQ-GL$_{1/2}$, SGL$_{1/2}$, SHSQRT-GL$_{1/2}$, and SHSQ-GL$_{1/2}$ achieved the sparsity more than $99$\%.
These results show the effectiveness of the proposed hierarchical group sparse regularizations.
Also, it is noticed that the group $L_{1/2}$ base regularizations are effective in increasing the sparseness. 

From these results, we can say that the parameters of the CNN are very redundant, and more than $90$\% of the weights are not necessary to achieve the classification accuracy of the baseline CNN without pruning.


We visualized the convolutional filters of each network after the training for the MNIST dataset with the structured sparse regularizations. 
Fig.~\ref{weight after training} shows the filters in the $1_{st}$ convolutional layer and the $2_{nd}$ convolutional layer for the baseline CNN and the networks trained with the neuron-wise HSQRT-GL and the feature-wise HSQRT-GL.

As shown in Fig.~\ref{weights of convolution in baseline network}, the weights of the trained filters without sparse regularization becomes active  at almost all locations and show various patterns. 
However, the filters trained with the hierarchical group sparse regularization are sparse in which many of the weights become almost zero, as shown in Fig.~\ref{weights of convolution in trained network with HSQRT-GL_neuron}  and Fig.~\ref{weights of convolution in trained network with HSQRT-GL_feature}.
Notably, only one filter remains active in the $1_{st}$ convolutional layer, and this filter works as a blurring filter.
Also, only a few filters are survived in the $2_{nd}$ convolutional layer.
These filters work as directional edge detection filters to the blurred input image processed by the filter in the $1_{st}$ convolution layer.
Interestingly, the effectiveness of the features obtained by the combinations of the directional edge filers and blurring is well known in character recognition, and there is a correspondence with the network trained with the structured sparse regularizations.  
We can get this structure automatically by training with sparse regularizations. 

Similar results are also obtained for the MNIST dataset and the Fashion-MNIST dataset by using the other structured sparse regularizations.
These results show that the network structure of a combination of the blurring filter and edge filters is fundamental for MNIST and Fashion-MNIST. 
It is interesting to consider the reason why this network structure is fundamental for gray image classification tasks.



Fig.~\ref{weights of convolution in trained network with HSQRT-GL_neuron} and Fig.~\ref{weights of convolution in trained network with HSQRT-GL_feature} show the visualizations of the filters obtained by the neuron-wise grouping with HSQRT-GL and the filters by feature-wise grouping with HSQRT-GL.
In the neuron-wise grouping, the weights connected to a output neuron are considered as a group, and the structured sparse regularization removes the neuron if the neuron is not necessary for the classification task.
On the other hand, the weights to a input neuron are considered as a group, and the structured sparse regularization removes the neuron if the neuron is not necessary for the classification task.

From the experiments, in both grouping methods, we found that the network trained with structured sparse regularization can remove unnecessary neurons enforcing the subset of the weights to be zero.
Thus, the trained filters with the neuron-wise grouping and the feature-wise grouping are similar.
In the following experiments, we show the results for the case of the feature-wise grouping.

\subsection{Hierarchical vs Non-Hierarchical}
\begin{table*}[htbp]
\centering
\caption{Results for AlexNet, ResNet18 and VGG11bn nets on CIFAR-10/100 and STL-10 dataset. Ave rank shows the average of the ranks of the sparsity. 
The best sparsity is shown in boldface.}
\label{tab_result}
\begin{tabular}{c||cccc||cccc||cc||c}\hline
Network & \multicolumn{4}{c||}{AlexNet} & \multicolumn{4}{c||}{ResNet18} & \multicolumn{2}{c||}{VGG11bn} &\multirow{3}{*}{\begin{tabular}{c}Ave\\rank\end{tabular}}\\ \cline{1-11}
Dataset & \multicolumn{2}{c}{CIFAR-10} & \multicolumn{2}{c||}{STL-10} & \multicolumn{2}{c}{CIFAR-10} & \multicolumn{2}{c||}{STL-10} & \multicolumn{2}{c||}{CIFAR-100} &\\ \cline{1-11}
Method & Accuracy & Sparsity & Accuracy & Sparsity & Accuracy & Sparsity & Accuracy & Sparsity & Accuracy & Sparsity &\\ \hline \hline
L2(Baseline) & 75.50\% & 0.50\% & 66.90\% & 1.52\% & 72.43\% & 1.59\% & 70.69\% & 1.13\% & 57.20\% & 1.09\% & -\\ 
L1 & 76.00\% & 5.53\% & 69.92\% & 12.34\% & 72.98\% & 72.82\% & 73.22\% & 73.75\% & 58.30\% & 39.32\% & 7.4\\
GL  & 76.03\% & 0.67\% & 69.53\% & 1.68\% & 73.44\% & 42.72\% & 74.66\% & 22.52\% & 58.59\% & 1.63\% & 14.4\\ 
ES & 75.70\% & 10.05\% & 69.39\% & 14.05\% & 72.79\% & 78.10\% & 72.84\% & 64.65\% & 58.87\% & 13.68\%  & 7.8\\ 
GL$_{1/2}$ & 75.76\% & 9.38\% & 69.33\% & 29.62\% & 73.29\% & 24.29\% & 71.62\% & 15.63\% & 57.44\% & 49.47\% & 8.8 \\ 
SGL & 75.91\% & 2.20\% & 67.36\% & 71.93\% & 73.48\% & 64.57\% & 72.12\% & 53.27\% & 58.56\% & 19.49\% & 9\\ 
SGL$_{1/2}$ & 75.55\% & 2.58\% & 67.04\% & 76.98\% & 73.16\% & 67.09\% & 71.56\% & 62.42\% & 58.49\% & 21.88\% & 7.4\\ 
CGES & 76.04\% & 2.46\% & 68.56\% & 30.21\% & 72.48\% & 88.14\% & 72.26\% & 10.01\% & 59.10\% & 1.29\% & 10\\ 
OICSR-GL & 76.23\% & 0.61\% & 69.54\% & 1.59\% & 73.69\% & 47.68\% & 75.02\% & 24.92\% & 59.03\% & 1.78\% & 14.2\\  \hline
HSQRT-GL & 76.02\% & 0.88\% & 69.49\% & 6.03\% & 73.03\% & 71.02\% & 70.94\% & 1.37\% & 58.91\% & 3.49\% & 13\\ 
HSQ-GL & 76.12\% & 3.16\% & 69.04\% & 23.11\% & 73.98\% & 67.96\% & 73.72\% & 40.52\% & 59.40\% & 7.35\%  & 9.8\\ 
HSQRT-ES & 75.80\% & 13.04\% & 70.10\% & 14.65\% & 73.02\% & 76.93\% & 71.40\% & 79.50\% & 58.46\% & 28.80\% & 5.6\\ 
HSQ-ES & 75.60\% & 10.25\% & 68.05\% & 17.66\% & 73.54\% & 58.49\% & 72.64\% & 25.22\% & 57.31\% & 14.97\%  & 9.4\\
HSQRT-GL$_{1/2}$ & 75.77\% & 5.63\% & 69.01\% & 28.02\% & 73.07\% & 83.54\% & 71.33\% & 9.66\% & 58.11\% & 43.61\% & 7.4 \\ 
HSQ-GL$_{1/2}$ & 75.54\% & \textbf{26.30\%} & 67.91\% & 63.20\% & 72.89\% & \textbf{89.05\%} & 70.71\% & \textbf{93.73\%} & 57.69\% & \textbf{62.79\%}  & \textbf{1.6}\\ 
SHSQRT-GL$_{1/2}$ & 75.80\% & 2.51\% & 67.06\% & \textbf{78.56\%} & 73.20\% & 68.11\% & 71.25\% & 67.39\% & 58.63\% & 21.79\% & 6.8\\ 
SHSQ-GL$_{1/2}$ & 75.67\% & 12.72\% & 69.10\% & 43.28\% & 72.80\% & 84.01\% & 72.67\% & 89.23\% & 57.72\% & 39.40\%  & 3.4\\ 
\hline
\end{tabular}
\end{table*}

\begin{figure*}[htbp]
\centering
\subfloat[filter-wise GL$_{1/2}$]{\includegraphics[scale=0.35]{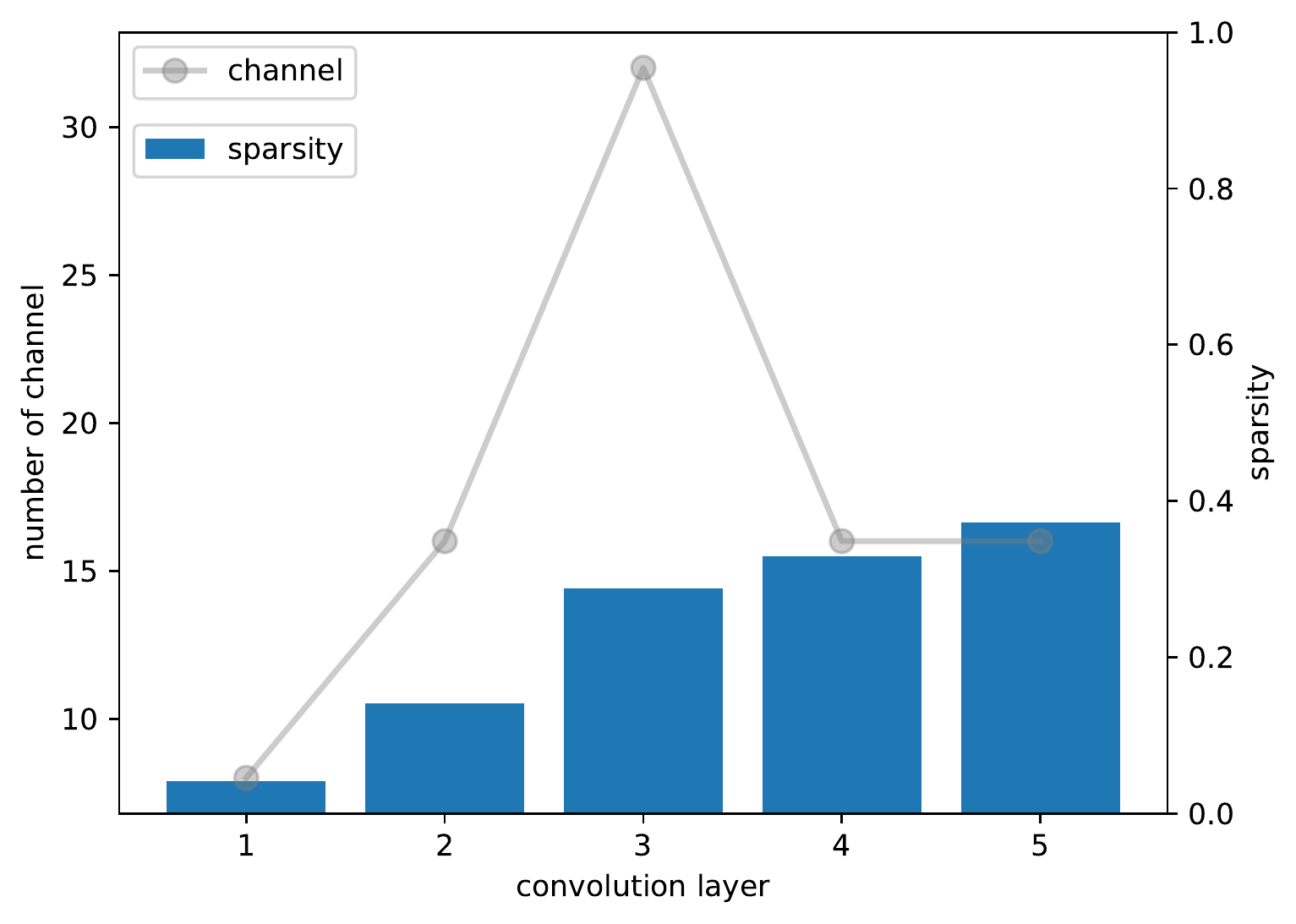}
\label{Sparsity Channel AlexNet on CIFAR10 with group L_1/2 regularization filter-wise grouping}}
\hfill
\subfloat[feature-wise GL$_{1/2}$]{\includegraphics[scale=0.35]{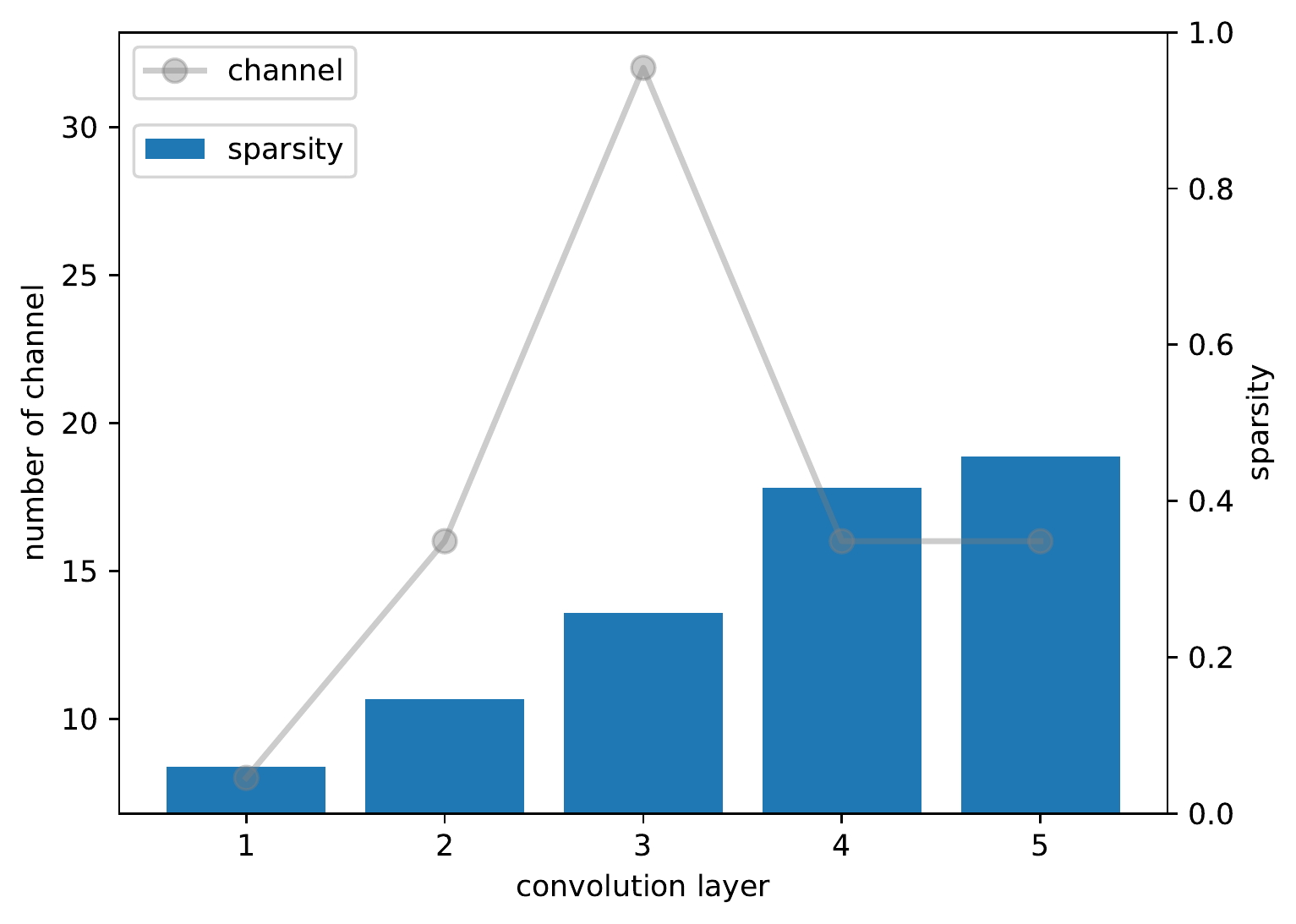}
\label{Sparsity Channel AlexNet on CIFAR10 with group L_1/2 regularization feature-wise grouping}}
\hfill
\subfloat[feature-wise HSQ-GL$_{1/2}$]{\includegraphics[scale=0.35]{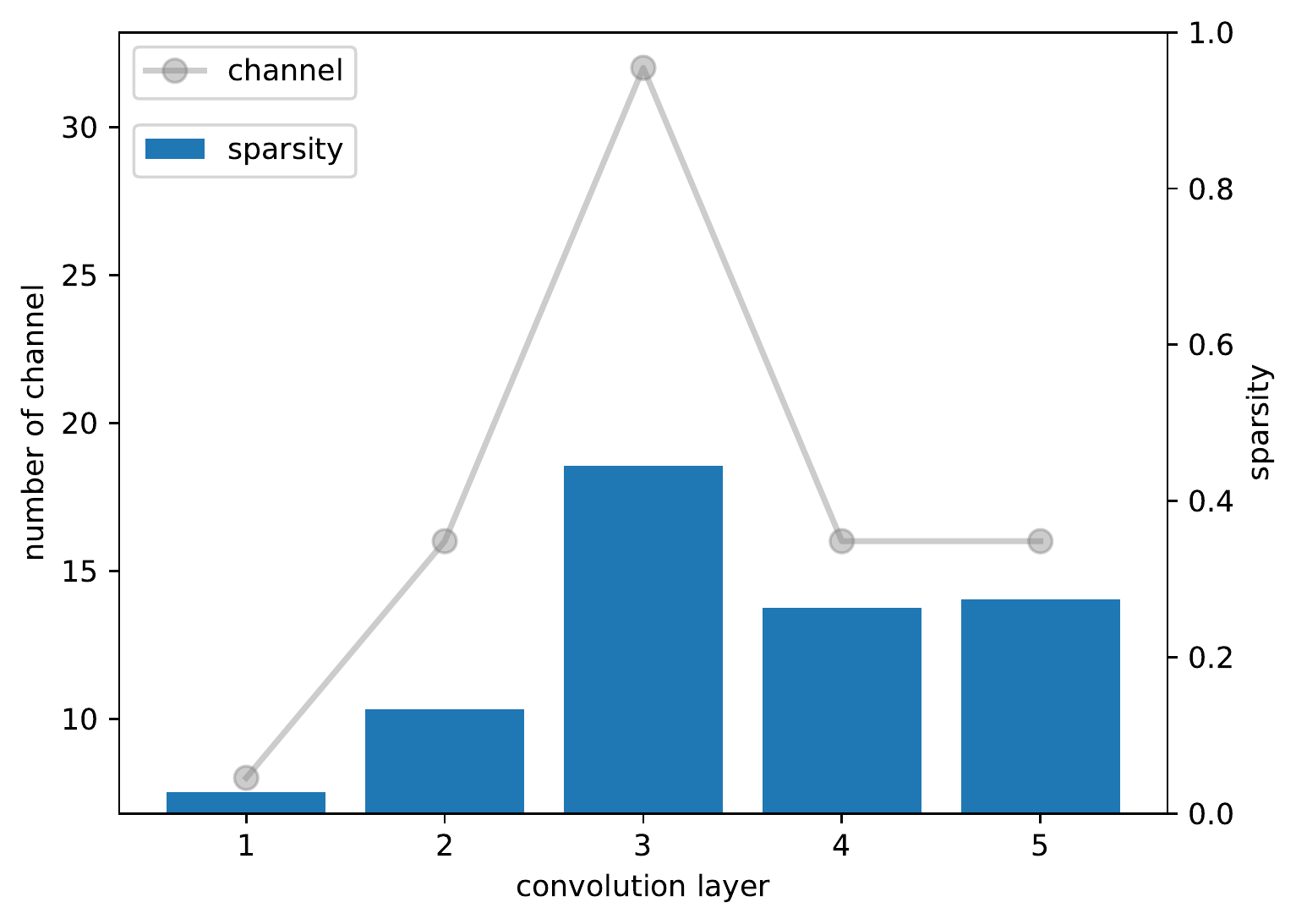}
\label{Sparsity Channel AlexNet on CIFAR10 with Hierarchical squared group L_1/2 regularization}}

\centering
\subfloat[filter-wise GL$_{1/2}$]{\includegraphics[scale=0.35]{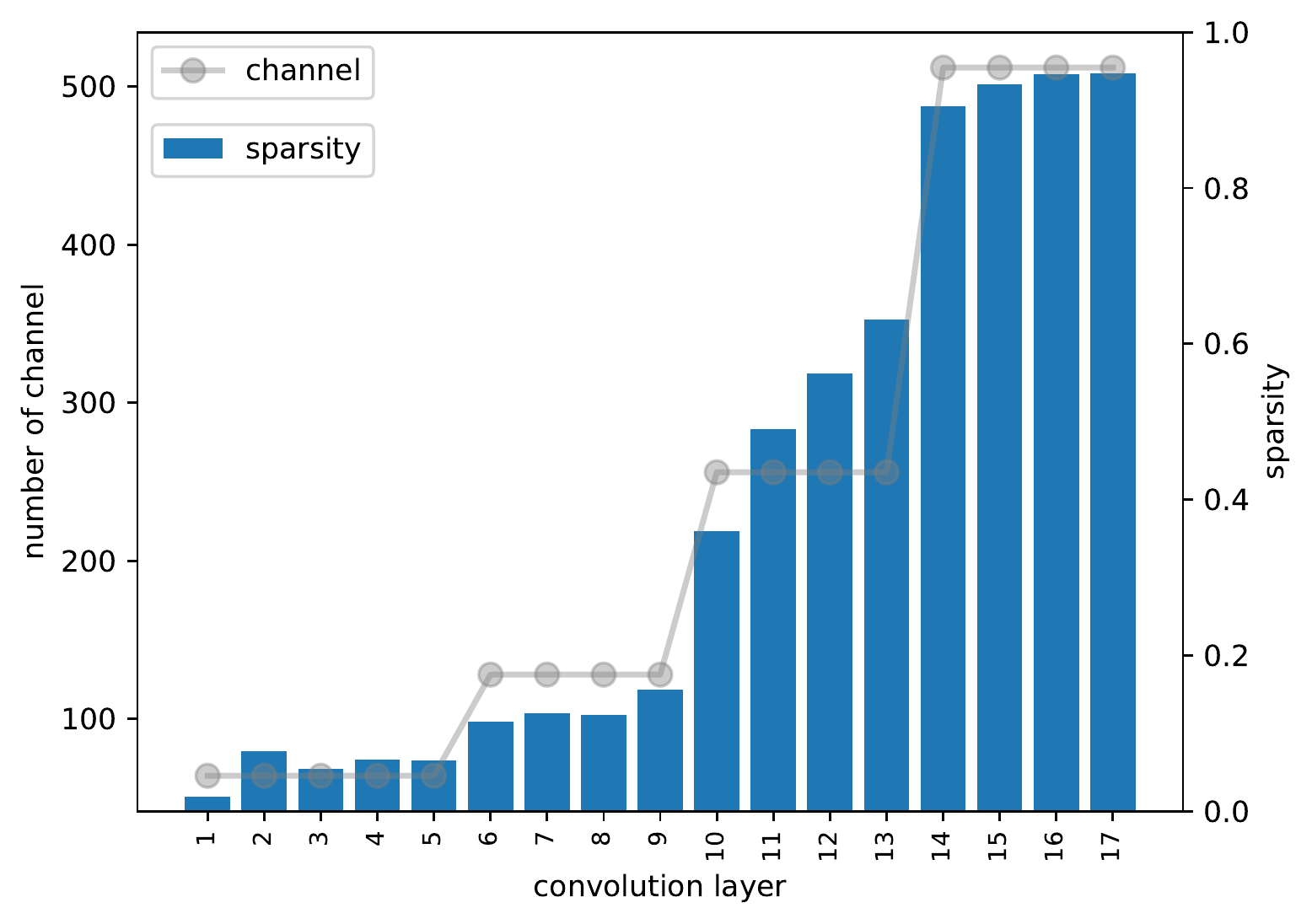}
\label{Sparsity Channel ResNet18 on STL10 with group L_1/2 regularization filter-wise grouping}}
\hfill
\subfloat[feature-wise GL$_{1/2}$]{\includegraphics[scale=0.35]{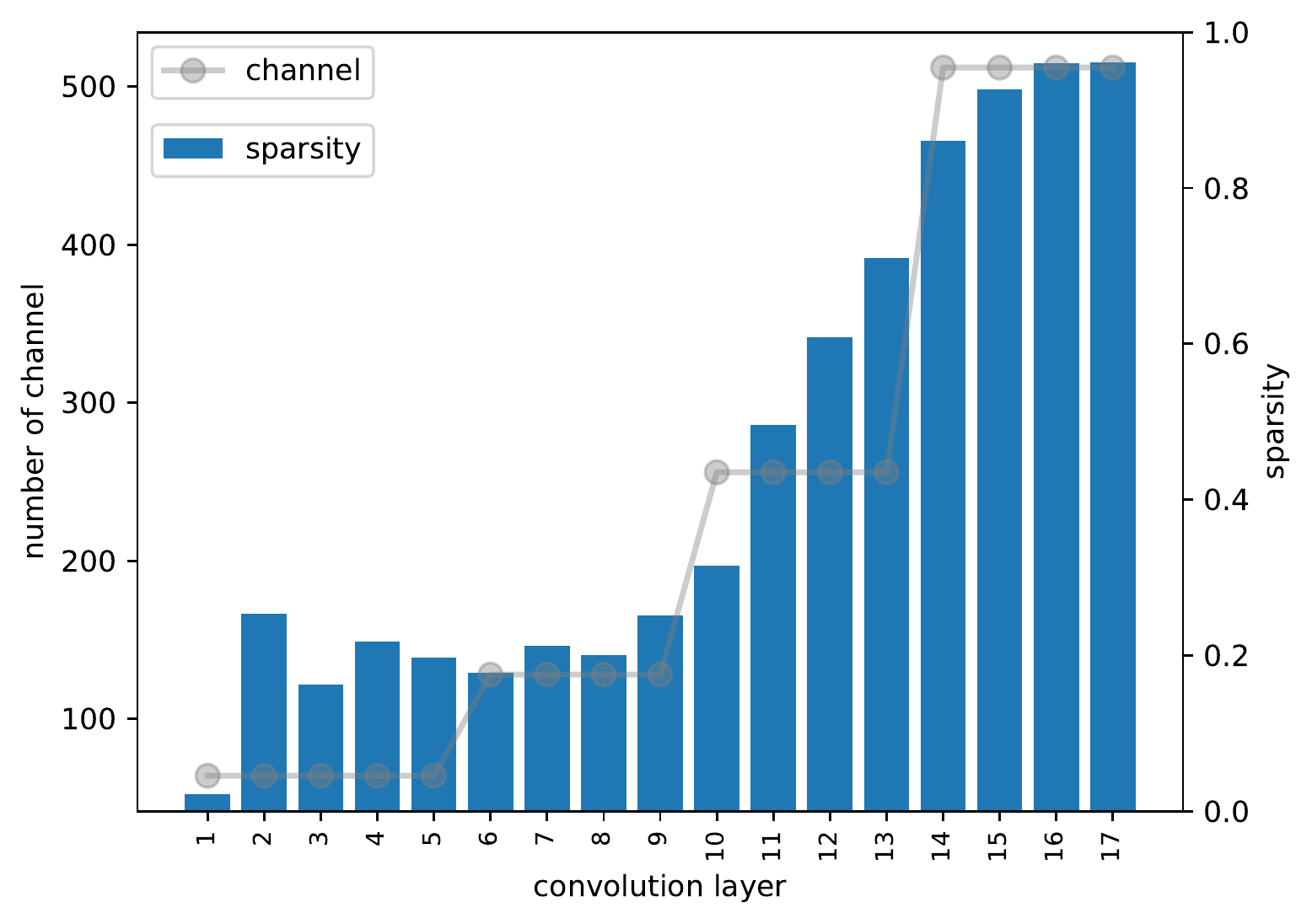}
\label{Sparsity Channel ResNet18 on STL10 with group L_1/2 regularization feature-wise grouping}}
\hfill
\subfloat[feature-wise HSQ-GL$_{1/2}$]{\includegraphics[scale=0.35]{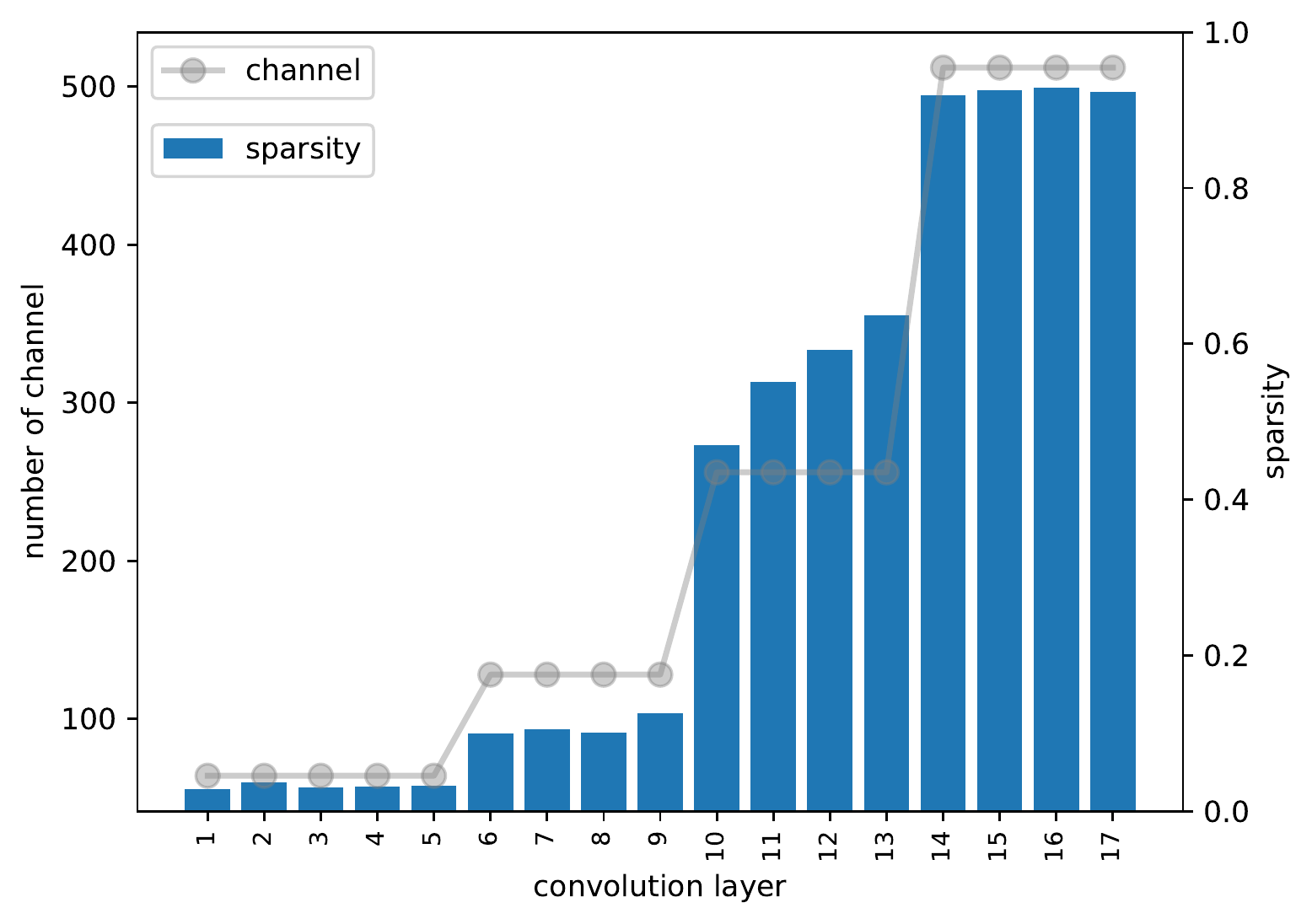}
\label{Sparsity Channel ResNet18 on STL10 with Hierarchical squared group L_1/2 regularization}}
\caption[]{The sparsity of each of the layers in the trained network with sparse regularization. 
\subref{Sparsity Channel AlexNet on CIFAR10 with group L_1/2 regularization filter-wise grouping}-\subref{Sparsity Channel AlexNet on CIFAR10 with Hierarchical squared group L_1/2 regularization} Sparsity of the trained AlexNet on CIFAR-10 with the group-wise GL$_{1/2}$.
The number of channels of the network are [ 8, 16, 32, 16, 16, 128, 10]. 
The sparsity of the trained AlexNet is around 30\%. 
The accuracy and sparsity are 78.23\%, 0.56\%.
\subref{Sparsity Channel ResNet18 on STL10 with group L_1/2 regularization filter-wise grouping}-\subref{Sparsity Channel ResNet18 on STL10 with Hierarchical squared group L_1/2 regularization} Sparsity the trained ResNet18 on STL-10 with group-wise GL$_{1/2}$.
The number of channels of the network are [ 64, 64, 64, 64, 64, 128, 128, 128, 128, 256, 256, 256, 256, 512, 512, 512, 512, 10]. 
The sparsity is around 80\%. 
The accuracy and sparsity are 75.97\%, 3.93\%.
\subref{Sparsity Channel AlexNet on CIFAR10 with group L_1/2 regularization filter-wise grouping} AlexNet on CIFAR-10 with the filter-wise GL$_{1/2}$ regularization, accuracy and sparsity are 78.99\%, 30.04\%.
\subref{Sparsity Channel AlexNet on CIFAR10 with group L_1/2 regularization feature-wise grouping}
AlexNet on CIFAR-10 with the feature-wise GL$_{1/2}$ regularization, accuracy and sparsity are 78.68\%, 33.67\%. 
\subref{Sparsity Channel AlexNet on CIFAR10 with Hierarchical squared group L_1/2 regularization}
AlexNet on CIFAR-10 with the feature-wise HSQ-GL$_{1/2}$ regularization, accuracy and sparsity are 79.25\%, 31.46\%. 
\subref{Sparsity Channel ResNet18 on STL10 with group L_1/2 regularization filter-wise grouping}
ResNet on STL-10 with the filter-wise GL$_{1/2}$ regularization, accuracy and sparsity are 78.48\%, 81.07\%. 
\subref{Sparsity Channel ResNet18 on STL10 with group L_1/2 regularization feature-wise grouping}
ResNet on STL-10 with the feature-wise GL$_{1/2}$ regularization, accuracy and sparsity are 78.36\%, 82.20\%. 
\subref{Sparsity Channel ResNet18 on STL10 with Hierarchical squared group L_1/2 regularization}
ResNet18 on STL-10 with the feature-wise HSQ-GL$_{1/2}$ regularization, accuracy and sparsity are 78.51\%, 80.85\%. 
}
\label{Sparsity at each convolution layers}
\end{figure*}



The training results of AlexNet, ResNet18, and VGG11bn with the sparse regularizations for the CIFAR-10, the CIFAR-100, and the STL-10 datasets are shown in Tab.~\ref{tab_result}. 

By comparing the proposed hierarchical group sparse regularizations (HSQRT-GL, HSQ-GL, HSQRT-ES, HSQ-ES, HSQRT-GL$_{1/2}$, and HSQ-GL$_{1/2}$) with the corresponding non-hierarchical group sparse regularization (GL, ES, and GL$_{1/2}$),  it is noticed that the networks trained with the hierarchical group sparse regularization are sparser than the non-hierarchical group sparse regularizations.
Especially, the hierarchical squared sparse regularizations can make the trained networks more sparse by keeping the accuracy the same as the baseline network. 


For all datasets and all network structures, HSQ-GL$_{1/2}$ shows the highest sparsity keeping high accuracy. 
These results show that HSQ-GL$_{1/2}$ is the most effective in terms of sparseness.





We also compare the ways of grouping with filter-wise grouping that the simplest grouping, feature-wise grouping that can consider the weight for the input neuron as a group and hierarchical feature-wise grouping that can consider the weight for the input neuron as a group and convolutional filter as a group in the same group.

Fig.~\ref{Sparsity at each convolution layers} shows the sparsity at each layer in AlexNet for CIFAR-10 and ResNet18 for STL-10 after training with the filter-wise GL$_{1/2}$, the feature-wise GL$_{1/2}$, and the feature-wise HSQ-GL$_{1/2}$. 

In AlexNet, the number of channels increases from the first layer to the third layer, and then it decreases at the forth layer.
The network trained with the filter-wise GL$_{1/2}$ or the feature-wise GL$_{1/2}$ regularizations becomes sparse only in the later layers.
On the other hand, the middle layers of the network trained with the feature-wise HSQ-GL$_{1/2}$ are also sparse.
These results suggest that we do not need to increase the number of channels in the later layers in AlexNet.


ResNet18 consists of four blocks of the layers. 
In each block, the number of channels is the same in each layer.
 The number of channels increases as the block becomes closer to the output layer.
The sparseness of each layer in the network trained with the filter-wise GL$_{1/2}$ regularization or the feature-wise GL$_{1/2}$ regularization increases as the layer is closer to the output.
Also, the sparseness of the layers in each block is different.
On the other hand, the sparseness of the layers is almost the same within each block when the network was trained with HSQ-GL$_{1/2}$ regularization. 
This result also shows that we do not need to increase the number of channels in ResNet18.

From these results, we think that the proposed hierarchical group sparse regularizations can effectively prune the unnecessary subsets of weights more adequately depending on the structure of the network and the number of channels.



\subsection{Comparison with the previous works}


From Tab.~\ref{tab_SimpleNet_on_MNIST_FashionMNIST} and Tab.~\ref{tab_result}, it is obvious that the sparseness of the network trained with the proposed hierarchical group sparse regularization criteria is higher than the previous sparse regularization criteria, including CGES and OICSR-GL.
It is noticed that higher sparsity keeping better accuracy than the base networks is obtained by the hierarchical group sparse regularization criteria, which use L1 regularization in the hierarchical grouping.
Especially HSQ-GL$_{1/2}$ gives the highest sparsity for almost all network architectures and datasets.







\bibliographystyle{unsrt}
\bibliography{root.bbl}

\begin{thebibliography}{10}

\bibitem{tibshirani1996regression}
Robert Tibshirani.
\newblock Regression shrinkage and selection via the lasso.
\newblock {\em Journal of the Royal Statistical Society: Series B
  (Methodological)}, 58(1):267--288, 1996.

\bibitem{zou2005regularization}
Hui Zou and Trevor Hastie.
\newblock Regularization and variable selection via the elastic net.
\newblock {\em Journal of the royal statistical society: series B (statistical
  methodology)}, 67(2):301--320, 2005.

\bibitem{yuan2006model}
Ming Yuan and Yi~Lin.
\newblock Model selection and estimation in regression with grouped variables.
\newblock {\em Journal of the Royal Statistical Society: Series B (Statistical
  Methodology)}, 68(1):49--67, 2006.

\bibitem{schmidt2010graphical}
Mark Schmidt.
\newblock Graphical model structure learning with $l_1$- regularization.
\newblock {\em University of British Columbia}, 2010.

\bibitem{Kim2010Tree}
Seyoung Kim and Eric~P. Xing.
\newblock Tree-guided group lasso for multi-task regression with structured
  sparsity.
\newblock In {\em Proceedings of the 27th International Conference on
  International Conference on Machine Learning}, page 543–550. Omnipress,
  2010.

\bibitem{friedman2010note}
Jerome Friedman, Trevor Hastie, and Robert Tibshirani.
\newblock A note on the group lasso and a sparse group lasso.
\newblock {\em arXiv preprint arXiv:1001.0736}, 2010.

\bibitem{simon2013sparse}
Noah Simon, Jerome Friedman, Trevor Hastie, and Robert Tibshirani.
\newblock A sparse-group lasso.
\newblock {\em Journal of computational and graphical statistics},
  22(2):231--245, 2013.

\bibitem{wen2016learning}
Wei Wen, Chunpeng Wu, Yandan Wang, Yiran Chen, and Hai Li.
\newblock Learning structured sparsity in deep neural networks.
\newblock In {\em Advances in neural information processing systems}, pages
  2074--2082, 2016.

\bibitem{alvarez2016learning}
Jose~M Alvarez and Mathieu Salzmann.
\newblock Learning the number of neurons in deep networks.
\newblock In {\em Advances in Neural Information Processing Systems}, pages
  2270--2278, 2016.

\bibitem{scardapane2017group}
Simone Scardapane, Danilo Comminiello, Amir Hussain, and Aurelio Uncini.
\newblock Group sparse regularization for deep neural networks.
\newblock {\em Neurocomputing}, 241:81--89, 2017.

\bibitem{zhou2010exclusive}
Yang Zhou, Rong Jin, and Steven Chu-Hong Hoi.
\newblock Exclusive lasso for multi-task feature selection.
\newblock In {\em Proceedings of the Thirteenth International Conference on
  Artificial Intelligence and Statistics}, pages 988--995, 2010.

\bibitem{kong2014exclusive}
Deguang Kong, Ryohei Fujimaki, Ji~Liu, Feiping Nie, and Chris Ding.
\newblock Exclusive feature learning on arbitrary structures via
  $\ell_{1,2}$-norm.
\newblock In {\em Advances in Neural Information Processing Systems}, pages
  1655--1663, 2014.

\bibitem{yoon2017combined}
Jaehong Yoon and Sung~Ju Hwang.
\newblock Combined group and exclusive sparsity for deep neural networks.
\newblock In {\em Proceedings of the 34th International Conference on Machine
  Learning-Volume 70}, pages 3958--3966. JMLR. org, 2017.

\bibitem{xu20101}
Zongben Xu, Hai Zhang, Yao Wang, XiangYu Chang, and Yong Liang.
\newblock $l_{1/2}$ regularization.
\newblock {\em Science China Information Sciences}, 53(6):1159--1169, 2010.

\bibitem{xu2012l_}
Zongben Xu, Xiangyu Chang, Fengmin Xu, and Hai Zhang.
\newblock $l_{1/2}$ regularization: A thresholding representation theory and a
  fast solver.
\newblock {\em IEEE Transactions on neural networks and learning systems},
  23(7):1013--1027, 2012.

\bibitem{zeng2014l_}
Jinshan Zeng, Shaobo Lin, Yao Wang, and Zongben Xu.
\newblock $l_{1/2}$ regularization: Convergence of iterative half thresholding
  algorithm.
\newblock {\em IEEE Transactions on Signal Processing}, 62(9):2317--2329, 2014.

\bibitem{wu2014batch}
Wei Wu, Qinwei Fan, Jacek~M Zurada, Jian Wang, Dakun Yang, and Yan Liu.
\newblock Batch gradient method with smoothing $l_{1/2}$ regularization for
  training of feedforward neural networks.
\newblock {\em Neural Networks}, 50:72--78, 2014.

\bibitem{fan2014convergence}
Qinwei Fan, Jacek~M Zurada, and Wei Wu.
\newblock Convergence of online gradient method for feedforward neural networks
  with smoothing $l_{1/2}$ regularization penalty.
\newblock {\em Neurocomputing}, 131:208--216, 2014.

\bibitem{li2018smooth}
Feng Li, Jacek~M Zurada, and Wei Wu.
\newblock Smooth group $l_{1/2}$ regularization for input layer of feedforward
  neural networks.
\newblock {\em Neurocomputing}, 314:109--119, 2018.

\bibitem{alemu2019group}
Habtamu~Zegeye Alemu, Junhong Zhao, Feng Li, and Wei Wu.
\newblock Group $l_{1/2}$ regularization for pruning hidden layer nodes of
  feedforward neural networks.
\newblock {\em IEEE Access}, 7:9540--9557, 2019.

\bibitem{li2019oicsr}
Jiashi Li, Qi~Qi, Jingyu Wang, Ce~Ge, Yujian Li, Zhangzhang Yue, and Haifeng
  Sun.
\newblock Oicsr: Out-in-channel sparsity regularization for compact deep neural
  networks.
\newblock In {\em Proceedings of the IEEE Conference on Computer Vision and
  Pattern Recognition}, pages 7046--7055, 2019.

\bibitem{ma2019transformed}
Rongrong Ma, Jianyu Miao, Lingfeng Niu, and Peng Zhang.
\newblock Transformed $\ell_{1}$ regularization for learning sparse deep neural
  networks.
\newblock {\em Neural Networks}, 119:286--298, 2019.

\bibitem{krizhevsky2012imagenet}
Alex Krizhevsky, Ilya Sutskever, and Geoffrey~E Hinton.
\newblock Imagenet classification with deep convolutional neural networks.
\newblock In {\em Advances in neural information processing systems}, pages
  1097--1105, 2012.

\bibitem{he2016deep}
Kaiming He, Xiangyu Zhang, Shaoqing Ren, and Jian Sun.
\newblock Deep residual learning for image recognition.
\newblock In {\em Proceedings of the IEEE conference on computer vision and
  pattern recognition}, pages 770--778, 2016.

\bibitem{simonyan2014very}
Karen Simonyan and Andrew Zisserman.
\newblock Very deep convolutional networks for large-scale image recognition.
\newblock {\em arXiv preprint arXiv: 1409.1556}, 2014.

\end{thebibliography}

\end{document}